\def\year{2022}\relax
\pdfoutput=1

\documentclass[letterpaper]{article} 
\usepackage{arxiv}  
\usepackage{times}  
\usepackage{helvet}  
\usepackage{courier}  
\usepackage[hyphens]{url}  
\usepackage{graphicx} 
\usepackage{natbib}  
\usepackage{caption} 
\DeclareCaptionStyle{ruled}{labelfont=normalfont,labelsep=colon,strut=off} 
\usepackage{chapterbib}

\usepackage{subcaption}
\usepackage{epsfig}
\usepackage{multirow}

\usepackage{amsthm}
\usepackage{amsfonts}

\usepackage{amsmath}
\usepackage{amssymb}
\usepackage{algorithm}
\usepackage{algorithmic}
\newtheorem{theorem}{Theorem}
\newtheorem{lemma}[theorem]{Lemma}
\newtheorem{remark}{Remark}
\newtheorem{assumption}{Assumption}
\usepackage{hyperref}
\def\eg{\emph{e.g., }}

\def\ie{\emph{i.e., }}
\def\etc{\emph{etc}}
\def\etal{{et al.}}

\usepackage{color}

\DeclareMathOperator*{\argmin}{argmin}  
\def\1{\textbf{1}}
\def\a{\textbf{a}}

\def\x{\textbf{x}}
\def\y{\textbf{y}}

\def\N{\mathcal{N}}

\def\R{\mathbb{R}}

\def\LL{\mathcal{L}}
\def\X{\mathcal{X}}
\def\Y{\mathcal{Y}}
\def\A{\mathcal{A}}

\def\ttheta{\boldsymbol{\theta}}

\def\tp{\text{TP}}
\def\tn{\text{TN}}
\def\fp{\text{FP}}
\def\fn{\text{FN}}

\usepackage{newfloat}
\usepackage{listings}
\floatstyle{ruled}
\newfloat{listing}{tb}{lst}{}
\floatname{listing}{Listing}
\usepackage{bibentry}
\usepackage{subfiles}

\title{Group-Aware Threshold Adaptation for Fair Classification}
\author{
    Taeuk jang\textsuperscript{\rm 1},
    Pengyi Shi\textsuperscript{\rm 2},
    Xiaoqian Wang\textsuperscript{\rm 1} \footnote{Corresponding author.}}
\affiliations{
    \textsuperscript{\rm 1}School of Electrical and Computer Engineering, Purdue University, West Lafayette, USA, 47907\\
    \textsuperscript{\rm 2}Krannert School of Management, Purdue University, West Lafayette, USA, 47907\\
    $\{$jang141, shi178, joywang$\}$@purdue.edu}
    
\begin{document}
\maketitle


\begin{abstract}
  The fairness in machine learning is getting increasing attention, as its applications in different fields continue to expand and diversify. To mitigate the discriminated model behaviors between different demographic groups, we introduce a novel post-processing method to optimize over multiple fairness constraints through group-aware threshold adaptation. We propose to learn adaptive classification thresholds for each demographic group by optimizing the confusion matrix estimated from the probability distribution of a classification model output. As we only need an estimated probability distribution of model output instead of the classification model structure, our post-processing model can be applied to a wide range of classification models and improve fairness in a model-agnostic manner and ensure privacy. This even allows us to post-process existing fairness methods to further improve the trade-off between accuracy and fairness. Moreover, our model has low computational cost. We provide rigorous theoretical analysis on the convergence of our optimization algorithm and the trade-off between accuracy and fairness of our method. Our method theoretically enables a better upper bound in near optimality than existing method under same condition. Experimental results demonstrate that our method outperforms state-of-the-art methods and obtains the result that is closest to the theoretical accuracy-fairness trade-off boundary.
\end{abstract}

\section{Introduction}
Machine learning is broadening its impact in various fields including credit analysis, job screening and \etc.
Consequently, importance of fairness in machine learning is emerging.
However, recent models have been found to behave differently between demographic groups in favorable predictions.
For example, it has been discovered that COMPAS, the criminal risk assessment software currently used to help pretrial release decisions, has biases between different races~\cite{dressel2018accuracy}. Specifically, blacks got higher risk scores predicted from the model than whites with similar profiles.
Therefore, discrimination truly exists and resolving it is critical as its direct and potential impact is growing tremendously.

However, obtaining fairness is not a trivial problem, {as the dataset itself will be biased when it is accumulated artificially.}
Simply modifying sensitive features (such as \emph{race}, \emph{gender}) from the data does not solve the bias, because there is indirect discrimination~\cite{pedreshi2008discrimination} caused by the feature relevance, which means sensitive information can be inferred from other features.

In order to alleviate discrimination from different perspectives, various quantitative measurements of group equity (Hardt \etal ~\shortcite{hardt2016equality}, Kleinberg \etal ~\shortcite{kleinberg2016inherent} Chouldechova \etal ~\shortcite{chouldechova2017fair})  have been proposed.
It has been proven that the pursuit of fairness is subject to a trade-off between fairness and accuracy (Liu \etal~\shortcite{liu2019implicit}, Kim \etal ~\shortcite{kim2020model}).

Moreover, Pleiss \etal~\shortcite{pleiss2017fairness} studied the trade-offs between fairness notions that cannot be satisfied at the same time.
Therefore, recent works (Feldman \etal ~\shortcite{feldman2015certifying}, Zhang \etal ~\shortcite{zhang2018mitigating}, Hardt \etal ~\shortcite{hardt2016equality}) usually target at a certain fairness notion.
However, these approaches suffer from the \textit{lack of flexibility}, \ie target fairness cannot be adjusted by the needs.
If the fairness constraints change under some circumstances, traditional fairness models need to be re-trained from scratch, which is computationally demanding and sometimes inapplicable due to model settings.

To overcome the limitations, we propose a novel post-processing method to improve fairness in a model-agnostic manner \ie we only need the prediction of an unknown model.
{GSTAR} (Group Specific Threshold Adaptation for faiR classification) model learns adaptive classification thresholds for each demographic group in classification task to improve the trade-off between fairness and accuracy. 
Given an existing classification model, GSTAR approximates the probability distribution of the model output and utilizes confusion matrix to quantify accuracy and fairness w.r.t. the group-aware classification thresholds. This allows us to: 1) prevent from burdening additional complexity or deteriorate the stability of the training process of the classifier; 2) integrate different fairness notions into one unified objective function; 3) easily adapt one pre-trained model to other fairness constraints.

We summarize our contributions of this paper as follows:
\begin{enumerate}
\item We propose a novel post-processing method, named GSTAR, which can learn group-aware thresholds to optimize the fairness-accuracy trade-off in classification. We empirically show that GSTAR outperforms state-of-the-art methods.
\item {With GSTAR, we can simultaneously optimize over multiple fairness constraints with a low computational cost. GSTAR does not require multiple iterations over data, instead, it takes \emph{at most} one pass of data in training for fast computation.}
\item {We conduct extensive rigorous theoretical analysis on our method, in terms of convergence analysis and fairness-accuracy trade-off. We theoretically prove a tighter upper bound of near optimality than existing method.}
\item{{We derive Pareto frontiers of our model for the fairness-accuracy trade-offs that contextualize the quality of fair classification.}}

\end{enumerate}

\section{Related Works}

In order to achieve group fairness, which quantifies the discrimination among different sensitive groups, a diverse notion of fairness has been introduced.
Equalized odds (Hardt \etal ~\shortcite{hardt2016equality}) enforce equality of true positive rates and false positive rates between different demographic groups. Pleiss \etal~\shortcite{pleiss2017fairness} relaxed equalized odds to satisfy the calibration.
Demographic parity or disparate impact~\cite{barocas2016big} suggests that a model is unbiased if the model prediction is independent of the protected attribute.
Among different fairness methods, post-processing techniques propose to improve fairness by modifying the output of a given classifier.
Hardt \etal~\shortcite{hardt2016equality} propose to ensure equalized odds by constraining the model output.
Kim \etal ~\shortcite{kim2020model} utilize confusion matrix and propose least-square accuracy-fairness optimization problem.
Kamiran \etal~\shortcite{kamiran2012decision} propose to give a favorable outcome to unprivileged and an unfavorable outcome to the privileged group when the confidence of the prediction is in a certain range. However, such \textit{static} confidence window keeps the same regardless of the demographic group and is determined by grid search, so it is less efficient.


Threshold adjustment (a.k.a. thresholding) was introduced to improve the performance of \textit{static} thresholds.
In the literature, Menon \etal~\shortcite{menon2018cost} prove that instance-dependent thresholding of the predictive probability function is the optimal classifier in cost-sensitive fairness measures. Also, when considering immediate utility, Corbett-Davies \etal~\shortcite{corbett2017algorithmic} show that optimal algorithm is achieved from group-specific threshold which is determined by group statistics.
However, to the best of our knowledge, the threshold adjustment approach has not been deeply studied that neither encompasses broad group fairness metrics nor describes an explicit method to achieve the threshold.

{Trade-off between fairness and accuracy} exists when we impose fairness constraint to a model.
Recent studies~\cite{chouldechova2017fair, zhao2019inherent} prove that models targeting at such fairness notions conform to an information theoretic lower bound on the joint error across different sensitive groups.
Therefore, our work presents a practical upper bound of the best achievable accuracy given the fairness constraints.

Here, our work is the most related to the post-processing methods (Hardt \etal ~\shortcite{hardt2016equality}, Kim \etal ~\shortcite{kim2020model}). 
{However, ours differ from theirs in several aspects. First, we theoretically prove that GSTAR achieves a better upper bound of near optimality than Hardt \etal~\shortcite{hardt2016equality} as we directly operate on ROC curve instead of linear intersections in Hardt \etal~\shortcite{hardt2016equality}.}
Also, GSTAR corrects the predicted label by the confidence of the prediction from a given model instead of randomly flipping the output to achieve equalized odds, which is more reliable in post-processing.
FACT (Kim \etal ~\shortcite{kim2020model}) utilizes a single point (static) from the classifier to be post-processed as a reference which does not fully utilize the classifier for the post-processing.
In contrast, by approximating the distribution of the continuous predicted logits, GSTAR model enables a larger feasible region than Kim \etal ~\shortcite{kim2020model} with a better fairness-accuracy trade-off. We validate the improvement in this trade-off via both theoretical and empirical results.
It is notable that these related methods can be considered as a special case of GSTAR.

\section{GSTAR for Fair Classification}
\subsection{Motivation}
Consider a binary classification problem with a binary sensitive feature, such that the sensitive feature ${A} \in \{0,1\}$ and label ${Y} \in \{0,1\}$.
In general, for a given data $X$, a binary classification model outputs an unnormalized logit $h(X)\in\R$ with the class label probability $R(X) = \sigma(h(X)) \in [0,1]$, where $\sigma$ is an activation function (\eg sigmoid function). It is not necessary to calculate $R$ in a classification model, \eg support vector machines directly use the positiveness/negativeness of logit $h(X)$ to determine classification outcome. 

For traditional models, we use a cut-off threshold $\theta_h = 0$ for $h(X)$ (\ie $\theta_R = \sigma(0) = 0.5$ for $R(X)$) in classification, such that the predicted label is determined by $\hat Y= \mathbb{I}\{h(X) \geq \theta_h\}$. {In the following context, unless otherwise mentioned, we use $\theta$ to refer to the threshold $\theta_h$ on logit $h$ since it is applicable to a wider range of classification models, and the corresponding threshold on label probability $\theta_R$ can be easily inferred from the threshold on logit $h$.
Traditional models use the same cut-off threshold $\theta$ for different demographic groups. However, since the distribution of logits $h$ in different demographic groups can be different, using the same threshold $\theta$ brings biased classification.

\begin{figure}[!t]
    \centering
    \includegraphics[width=0.9\linewidth, height=0.75\linewidth]{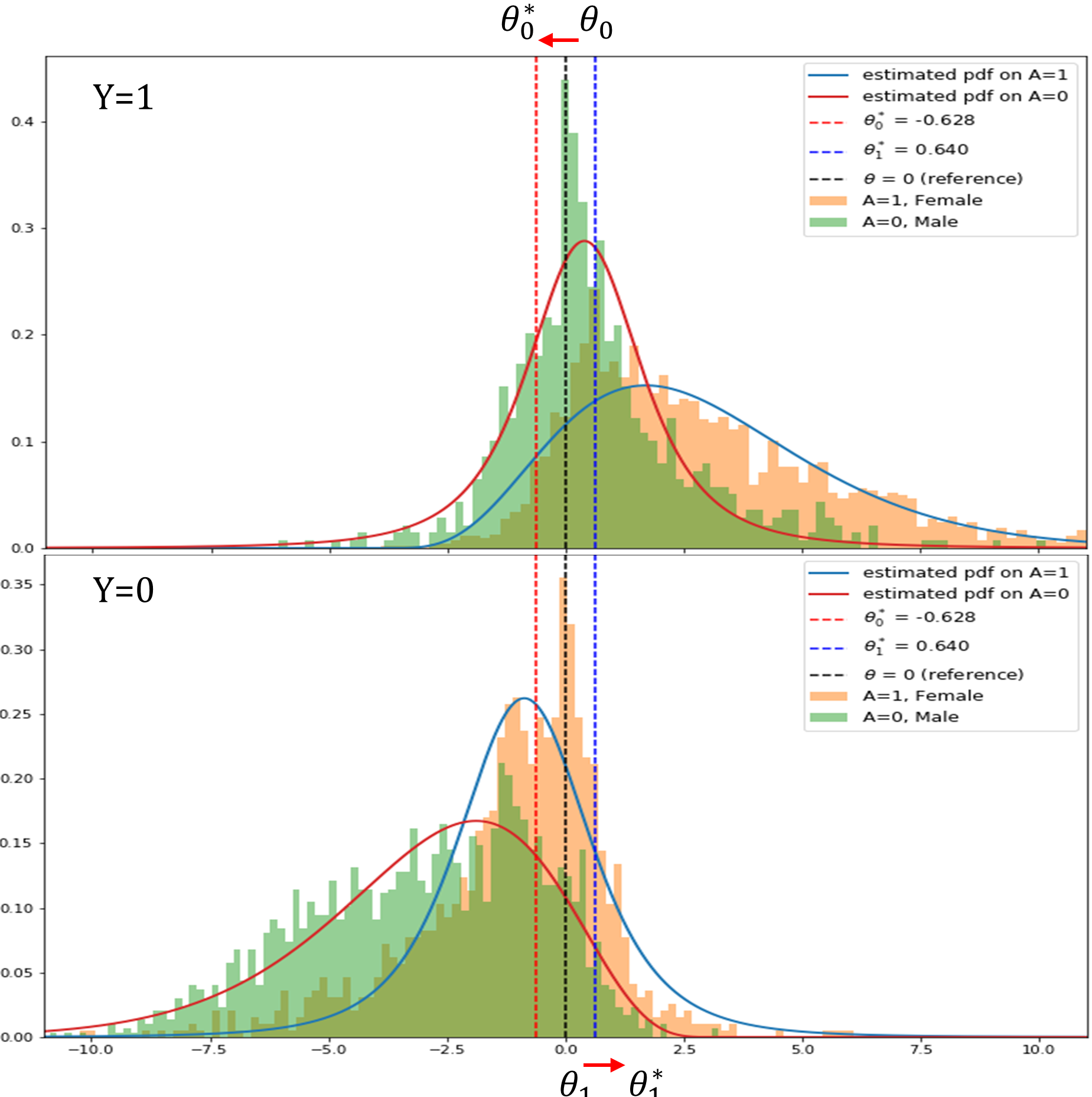}
    \caption{
    {Histograms of logit $h$ distribution} from logistic regression on CelebA data, where $\theta$ is the threshold to assign predicted label based on $h$. 
    The top and bottom plot is for positive samples ($Y=1$, attractive), and negative samples ($Y=0$, unattractive).
    Bars represent the distributions of logit $h$ of sensitive groups, and curves are estimated probability density functions of logit $h$ of sensitive groups as in the legend.
    $\theta = 0$ (black dashed line) is the default classification thresholds.
    The default thresholds result in biased prediction towards the unprivileged group ($A=0$) due to the different logit $h$ distributions in different sensitive groups. $(\theta^*_0,\theta^*_1)$ (colored dashed line) are group-aware thresholds for each sensitive group achieved by GSTAR.}
    \label{fig:illust}
\end{figure}

In~\autoref{fig:illust}, we show a real-world example of image classification on CelebA dataset with ResNet50~\cite{he2016deep} to show
{that the default setting of classification thresholds affects both accuracy and fairness in classification.}
The goal is to predict the image of a person is whether attractive or not, and consider sensitive attribute as gender. 
This can be generalized to different sensitive attributes, \eg age or race~\cite{lokhande2020fairalm}.
We can observe an obvious difference in the distribution of logit $h$ between two gender groups. 
If we use a unified classification threshold $\theta_1 = \theta_0 = 0$,
it naturally brings a difference in the true positive rate and true negative rate between two gender groups, thus it behaves as a biased classification.
Instead, we observe that the optimal group-specific threshold obtained from GSTAR ($\theta_1^*>\theta_1$, and $\theta_0^*<\theta_0$) can adapt to such discrepancy in distribution between two demographic groups to improve both fairness and accuracy.}

\subsection{Group-Aware Classification Thresholds}

Given an existing classification model and a sensitive attribute $a$, we can denote true positive rate ($\text{TP}_a$), false positive rate ($\text{FP}_a$), true negative rate ($\text{TN}_a$), and false negative rate ($\text{FN}_a$) in the confusion matrix. 
Most fairness notions can be represented with entries in the confusion matrix. For instance, Equal Opportunity (EOp) (Hardt \etal ~\shortcite{hardt2016equality}) requires $TP_0 = TP_1,$ and Demographic Parity (DP)~\cite{barocas2016big} requires
\[
\frac{TP_1 n_{11} + FP_1 n_{01}}{N_1} = \frac{TP_0 n_{10} + FP_0 n_{00}}{N_0},
\]
where $n_{ya}$ denotes the number of samples in the subset $\{Y = y, A = a\}$, $N_a = \sum_y n_{ya}$ denotes the number of samples in $\{Y = y\}$, and $N=\sum_{y,a}n_{ya}$ is the total number of samples.

Consider the group-aware classification threshold $\boldsymbol{\theta}=(\theta_1, \theta_0)^\mathsf{T}$, where $\theta_a$ is the classification threshold for sensitive group $A=a$. We can formulate the entries in the confusion matrix w.r.t. $\boldsymbol{\theta}$ as below:
\begin{eqnarray}
\begin{aligned}
\label{eq:the-con}
    &\text{TP}_a(\theta_a) \approx 1- \int_{-\infty}^{\theta_a}f_{1a}(x)dx,
    &\text{FN}_a(\theta_a) \approx 1 - \text{TP}_a(\theta_a)\\
    &\text{FP}_a(\theta_a) \approx 1- \int_{-\infty}^{\theta_a}f_{0a}(x)dx,
    &\text{TN}_a(\theta_a) \approx 1 - \text{FP}_a(\theta_a)
\end{aligned}
\end{eqnarray}
where $f_{ya}(x)$ is an estimated probability density function of the distribution of output logit $h$ in the subset $\{Y = y, A = a\}$.


Then, we formulate the fairness-constrained classification problem with the objective of minimizing classification error into a least-squared optimization problem. 
We denote our objective function as $\LL(\ttheta)$ which consists of the performance loss $\LL_{per}(\ttheta)$ and fairness loss $\LL_{fair}(\ttheta)$ that are represented with the entries of the confusion matrix.
In other words, our goal is to minimize the objective function $\LL(\ttheta)$ as below:
\begin{equation}
\label{eq:obj}
\LL(\ttheta) = \LL_{per}(\ttheta) + \lambda \LL_{fair}(\ttheta),
\end{equation}
where $\lambda$ is a hyperparameter that determines how much fairness is enforced in the optimization.
The performance error $\LL_{per}(\ttheta)$ can be written as
\begin{multline*}
     \LL_{per}(\ttheta) = \Big(\frac{n_{01}}{N}\fp_1(\theta_1) + \frac{n_{11}}{N}\fn_1(\theta_1) \\
                        + \frac{n_{00}}{N}\fp_0(\theta_0) + \frac{n_{10}}{N}\fn_0(\theta_0)\Big)^2. 
\end{multline*} 

{As for $\LL_{fair}(\ttheta)$, it can be formulated to any fairness metrics that are expressible with confusion matrix.
For instance, when we impose EOp ($\tp_1 = \tp_0)$ and predictive equality (PE) ($\fp_1 = \fp_0)$~\cite{chouldechova2017fair}, we can get the corresponding $\LL_{fair}(\ttheta)$ by summing over the least squared form of each constraint.
Also, satisfying EOp and PP is equivalent to satisfying Equalized Odds (EOd)~\cite{hardt2016equality}, 
This can be formulated in our $\LL_{fair}$ as 
\begin{eqnarray}
\label{eq:perfect_eod}
\begin{aligned}
&\LL_{fair}^{EOd}(\ttheta) =\LL_{fair}^{EOp}(\ttheta) + \LL_{fair}^{PP}(\ttheta) \\
= \big( TP_1(\theta_1) &- TP_0(\theta_0) \big)^2 + \big( FP_1(\theta_1) - FP_0(\theta_0) \big)^2.
\end{aligned}
\end{eqnarray}
Note that a lower $\LL_{fair}$ value indicates a fairer threshold. When $\LL_{fair}^{EOD}(\ttheta)= 0 $, we can interpret as the $\ttheta$ satisfies the perfect EOd fairness.
Similar to~\eqref{eq:perfect_eod}, we can enforce multiple fairness constraints by summing over the least squared of each metric with different weight constant $\lambda$ to each fairness constraints if needed.

}
Also, it is notable that compared to FACT (Kim \etal ~\shortcite{kim2020model}) that enforces fairness through confusion tensor, our formulation of fairness in $\LL_{fair}(\ttheta)$ represents a direct notion of fairness metrics and improves the measures that allows us to achieve better performance and Pareto frontiers that is shown in Section~\ref{sec:trade-off} and~\autoref{fig:frontier}.
For example, FACT integrates multiple constraints as a weighted sum with the weights being the number of samples in each class. In this expression, the imbalance between the two fairness criteria will grow as the degree of imbalance in the data increases. 
In contrast, our formulation expresses the constraints as the exact notion of each metric that is not biased by the statistics of the datset and we observe {improved Pareto frontier as in~\autoref{fig:frontier}}.

\subsection{Optimization of GSTAR}
Our GSTAR objective in \eqref{eq:obj} lies in the family of Nonlinear Least Squares Problem (NLSP)~\cite{gratton2007approximate}. To optimize objective \eqref{eq:obj} and find the threshold $\boldsymbol{\theta}$, we adopt the Gaussian-Netwon optimization method~\cite{gratton2007approximate}.
Here we take EOp constraint as an example to show the alternating optimization steps, then $\LL_{fair}(\ttheta)$ can be written as
\begin{equation}
    \LL_{fair}^{EOp}(\ttheta) = \left(\tp_1(\theta_1) - \tp_0(\theta_0)\right)^2.
\end{equation}

To solve NLSP with the Gauss-Newton method,  first convert the nonlinear optimization problem to a linear least square problem using Taylor expansion. That is, the parameter values are calculated in an iterative fashion with 
\begin{equation}
\theta_{a} \approx \theta_{a}^{k+1} = \theta_{a}^{k} +\Delta_a,    
\end{equation}
in the $k$-th iteration number, with the vector of increments $\Delta=\{\Delta_a\} = \{ \theta_{a}^{k+1} - \theta_{a}^{k} \}$ (also known as the shift vector).

We rewrite our objective function as a real vector function $r(\boldsymbol{\theta}) = \big( r_1(\boldsymbol{\theta}), r_2(\boldsymbol{\theta}) \big) = (\LL_{per}, \lambda \LL_{fair})$.
We linearize each component in the loss function to a first-order Taylor polynomial expansion as 
\begin{equation}
r_i (\boldsymbol{\theta}) \approx r_i(\boldsymbol{\theta}^k) + \sum_{a} \frac{\partial r_i(\boldsymbol{\theta}^k)}{\partial\theta_a} \Delta_a 
\label{eq:Taylor}
\end{equation}
with $\boldsymbol{\theta}^k =(\theta_0^{k}, \theta_1^k)$. Plugging this linearized equation into the objective function, we get the usual least square problem. Then, the optimal solution can be obtained as 
\begin{equation}
\Delta = - (J^T J)^{-1} J^T f(\boldsymbol{\theta}^k), 
\label{eq:NLSP-solution}
\end{equation}
where $J=\{J_{ia}\}$ with $J_{ia} = \{\frac{\partial r_i(\boldsymbol{\theta})}{\partial\theta_a}\}$ is the Jacobian. 
Each entry of the jacobian can be expressed with linear combination of pdf and cdf of $f_{ya}$ for $i,a,y\in\{0,1\}$.
we can finalize the alternating optimization as
\begin{equation}
    \theta_0^{\tau} = \theta_0^{\tau-1} + \Delta_0^{\tau}, \quad  \theta_1^{\tau} = \theta_1^{\tau-1} + \Delta_1^{\tau}.
\end{equation}
{It is notable that in each iteration we derive the optimal update step $\Delta_a$, which eliminates the burden of tuning hyperparameter (such as learning rate) in iterative algorithm.}
See the supplementary for detailed optimization process.

The alternating optimization of GSTAR model is of low computational cost. 
We take at most one pass of the data for learning the estimated probability density functions $f_{ya}$ in~\eqref{eq:the-con} (we do not even need to traverse the data if the parameters (such mean and variance in Gaussian distribution) for the estimated probability density functions $f_{ya}$ can be provided). 
The optimization of $\ttheta$ with alternating optimization is efficient since we only need $f_{ya}$. 
Therefore, we need a constant time for each update.
Overall, the time complexity of GSTAR is $O(n+T)$, where $n$ is the number of samples, and $T$ is the number of iterations in alternating optimization.

Besides, if a unified threshold is necessary~\cite{corbett2017algorithmic}, \ie $\theta_1 = \theta_0$, the optimization algorithm also applies and we only have one scalar variable in~\eqref{eq:obj}.
When we have a unified threshold, we do not require sensitive information in the testing phase that we can conform more strict privacy regulations than group-aware thresholding. However, we have to sacrifice both fairness and accuracy as the thresholding is less flexible.

\subsection{Theoretical Analysis}
\subsubsection{Upper Bounds on FPR/FNR Gap Between Groups}
We first state the assumptions we need to make for Theorem \ref{thm1} and \ref{thm2}.
\begin{assumption}
\label{assump-classify}
For any given classier $h$ and its induced PDF $f_{ya}$ and CDF $F_{ya}$, we assume the following holds:  
\begin{itemize}
    \item The PDF $f_{ya}(x)$ is uniformly bounded, i.e., there is an $\hat{f}_{ya}(x) = \max_x f_{ya}(x)$. 
    \item The inverse CDF $F^{-1}_{ya}(x)$ is Lipschitz continuous with Lipschitz constant $M_{ya}$. 
    \item The difference in the CDF between two groups is uniformly bounded, i.e., 
    $$ | F_{y1}(x) - F_{y0}(x) | \leq u_y, ~~\forall x. $$
\end{itemize}
\end{assumption}

\begin{theorem}
\label{thm1}
For any given classifier that satisfies Assumption~\ref{assump-classify} and any given pair of thresholds $(\theta_0, \theta_1)$ that satisfies the perfect EOp condition, the gap between false-positive rates of the two group is upper bounded by
\begin{equation}
|\epsilon_1| = \big| \fp_0(\theta_0)
- \fp_1(\theta_1) \big| 
\leq u_0 + C_1 u_1 ,  
\end{equation}
where $C_1 = \hat{f}_{01} M_{10}$. 
\end{theorem}
\begin{theorem}
\label{thm2}
For any given classifier that satisfies Assumption~\ref{assump-classify} and any given pair of thresholds $(\theta_0, \theta_1)$ that satisfies the perfect PE condition, the gap between false-negative rates of the two group is upper bounded by
\begin{equation}
|\epsilon_2| = \big| \fn_0(\theta_0) - \fn_1(\theta_1) \big| 
\leq u_1 + C_0 u_ 0,  
\end{equation}
where $C_0 = \hat{f}_{11} M_{00}$. 
\end{theorem}

Theorem \ref{thm1} and \ref{thm2} characterize the upper bound of false positive/negative rate gap between two groups when the false negative/positive rate gap is 0. At the same time, it captures the upper bound of additional accuracy loss due to the two different thresholds for different groups under a perfect fairness (EOp or EP) condition. 

\subsubsection{Trade-off between Accuracy and Fairness}
Now we prove a theorem to characterize the trade-off between accuracy and fairness.
Let $\theta^*_a = \argmin_{\theta_a}  \LL_{per}(\theta_a),$ and its perturbed value $\tilde \theta_a$ as
\begin{equation}
\begin{split}
  |\fn_1({\theta_1}^*) - \fn_1({\tilde{\theta}_1})|
\leq \gamma/2, \\
|\fn_0({\theta_0}^*) - \fn_0({\tilde{\theta}_0})|
\leq \gamma/2,
\end{split}
\label{eq:perturb-assump}
\end{equation}
for some perturbation coefficient $\gamma$. Then for optimal perturbed version $\tilde \theta^*_a = \argmin_{\tilde \theta_a}  \LL_{per}(\tilde \theta_a)$, we state the theorem below:
\begin{theorem}
\label{thm3}
Under Assumption~\ref{assump-classify} and condition~\eqref{eq:perturb-assump},
\begin{eqnarray*}
\LL_{per}(\theta_1^*) - \LL_{per}(\tilde{\theta}_1^*)
&\leq& C\gamma, 
\end{eqnarray*}
where 
$$ C = 2L^* \bigg( \frac{r_1}{2} + r_0 \frac{\hat{f}_{01}M_{11}}{2}
+ \frac{n_{00}}{N}\Big(\hat{f}_{00}M_{10}  + \frac{\hat{\epsilon_1^\prime} M_{11}}{2}\Big) + \frac{n_{10}}{N}
\bigg) $$
and $\hat{\epsilon}_1^\prime = \max \tilde{\epsilon}_1^\prime$ is the maximum of the derivative of $\tilde{\epsilon}_1$.  
\end{theorem}

Theorem \ref{thm3} quantifies the decrease in accuracy loss (\ie the improvement in accuracy) when we allow a gap of true positive rates between two groups (i.e., relaxation from the perfect EOp condition).

\subsubsection{Convergence Analysis of GSTAR}
Our objective function and the optimization solution algorithm belong to the family of Gauss-Newton algorithm.
Given the assumptions A1 and A2 below,
\begin{itemize}
    \item A1. There exists $\theta^*$ such that $J^T(\theta^*)r(\theta^*) = 0$,
    \item A2. The Jacobian at $\theta^*$ has full rank,
\end{itemize}
we state the following theorem of convergence:
\begin{theorem}
Assume that the estimated density function $f(\cdot)$ satisfy assumptions A1 and A2. Further, $f(\cdot)$ satisfies that 
\[
||Q(\theta^k)(J^T J)^{-1}(\theta^k)||_2 \leq \eta 
\]
for some constant $\eta\in[0,1)$ for each iteration $k$, where $Q(\theta)$ denotes the second order terms $\sum_i r_i(\theta) \nabla^2 r_i(\theta)$. Then as long as the initial solution is sufficiently close to the true optimal with $||\theta^0 -\theta^*||_2 \leq \epsilon$, the sequence of Gauss-Newton iterates $\{\theta^k\}$ converges to $\theta^*$. 
\end{theorem}

\subsubsection{Near Optimality of GSTAR}
Following the proof of Theorem 5.6 of Hardt \etal ~\shortcite{hardt2016equality}, we provide the following near optimality theorem for our GSTAR model.
\begin{theorem}
With a bounded loss function $\ell$ and a given estimated density function $h(x)$, let $\hat{R}_h \in [0,1]$ be the induced random variable from the density $h(x)$. Then the equalized odds predictor $\hat{Y}_h$ derived from $(\hat{R}_h, A)$ using the method in our paper can achieve near optimality in the following sense:
$$\mathbb{E}[\ell(\hat{Y}_h, Y)] \leq \mathbb{E}[\ell(Y^*, Y)] + 2 d_K(\hat{R}_h, R^*).$$
Here, $Y$ is the true label, $Y^*$ is the optimal equalized odds predictor derived from the Bayes optimal regressor $R^*$ as given in Hardt et al. \cite{hardt2016equality}, and $d_K(\hat{R}_h, R^*)$ is the conditional Kolmogorov distance.
\label{thm5}
\end{theorem}
Theorem \ref{thm5} provides that GSTAR has tighter bound of near optimality than Hardt \etal~\shortcite{hardt2016equality} under the same condition.
See the supplementary for the proof of Theorem \ref{thm1} $\sim$ \ref{thm5}.

\section{Experiments} \label{sec:ex}
In this section, we validate GSTAR model on four well-known fairness datasets and compare with other state-of-the-art methods.

\subsection{Experimental Setup}
\label{subsec:setup}
\begin{figure}[!t]
    \centering
    \begin{minipage}[t]{0.5\textwidth}
		\centering
		\includegraphics[width=1.0\textwidth, height=0.43\textwidth]{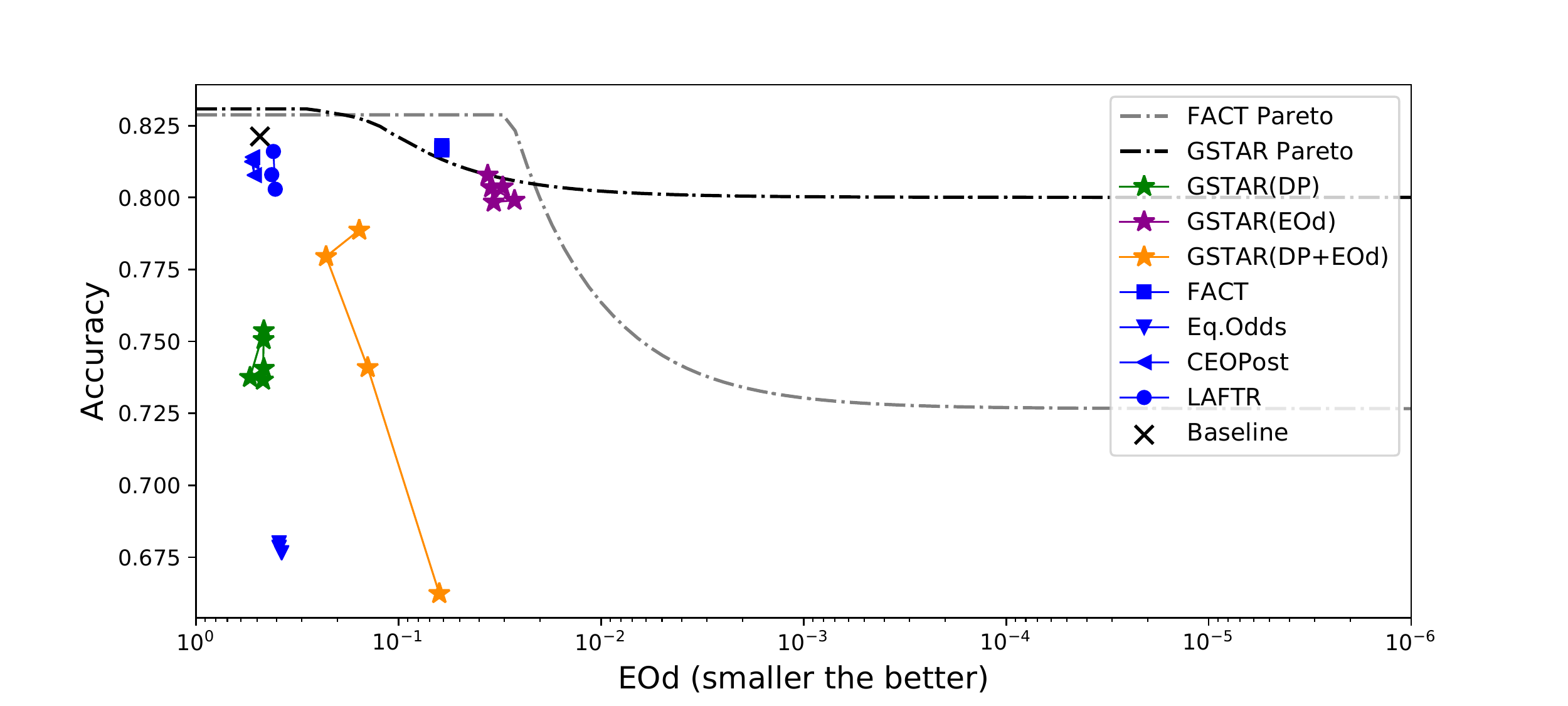}
         \subcaption{CelebA Dataset}
         \label{fig:front_celeba}
	\end{minipage}
    \begin{minipage}[t]{0.5\textwidth}
		\centering
         \includegraphics[width=1.0\textwidth, height=0.43\textwidth]{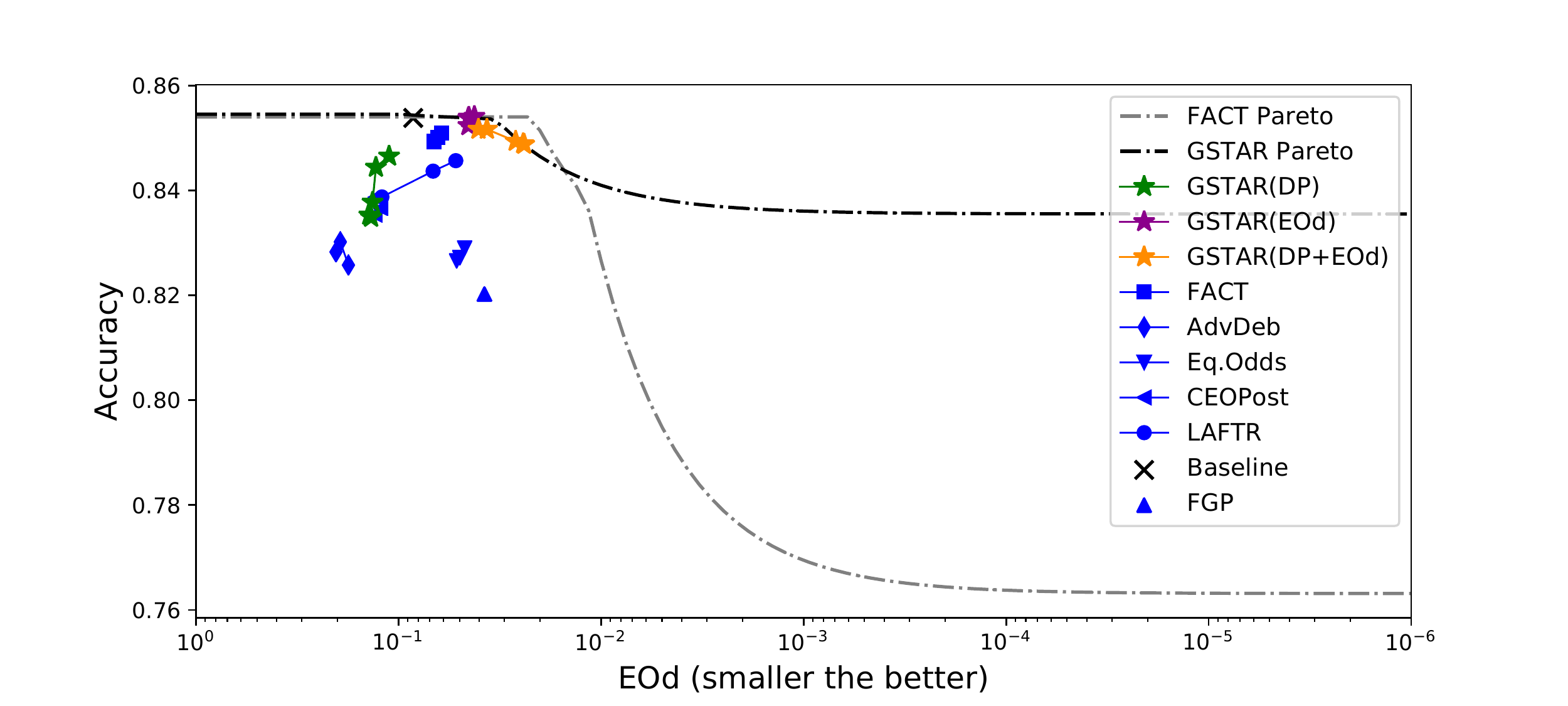}
         \subcaption{Adult Dataset}
         \label{fig:front_adult}
	\end{minipage}
    \begin{minipage}[t]{0.5\textwidth}
		\centering
         \includegraphics[width=1.0\textwidth, height=0.43\textwidth]{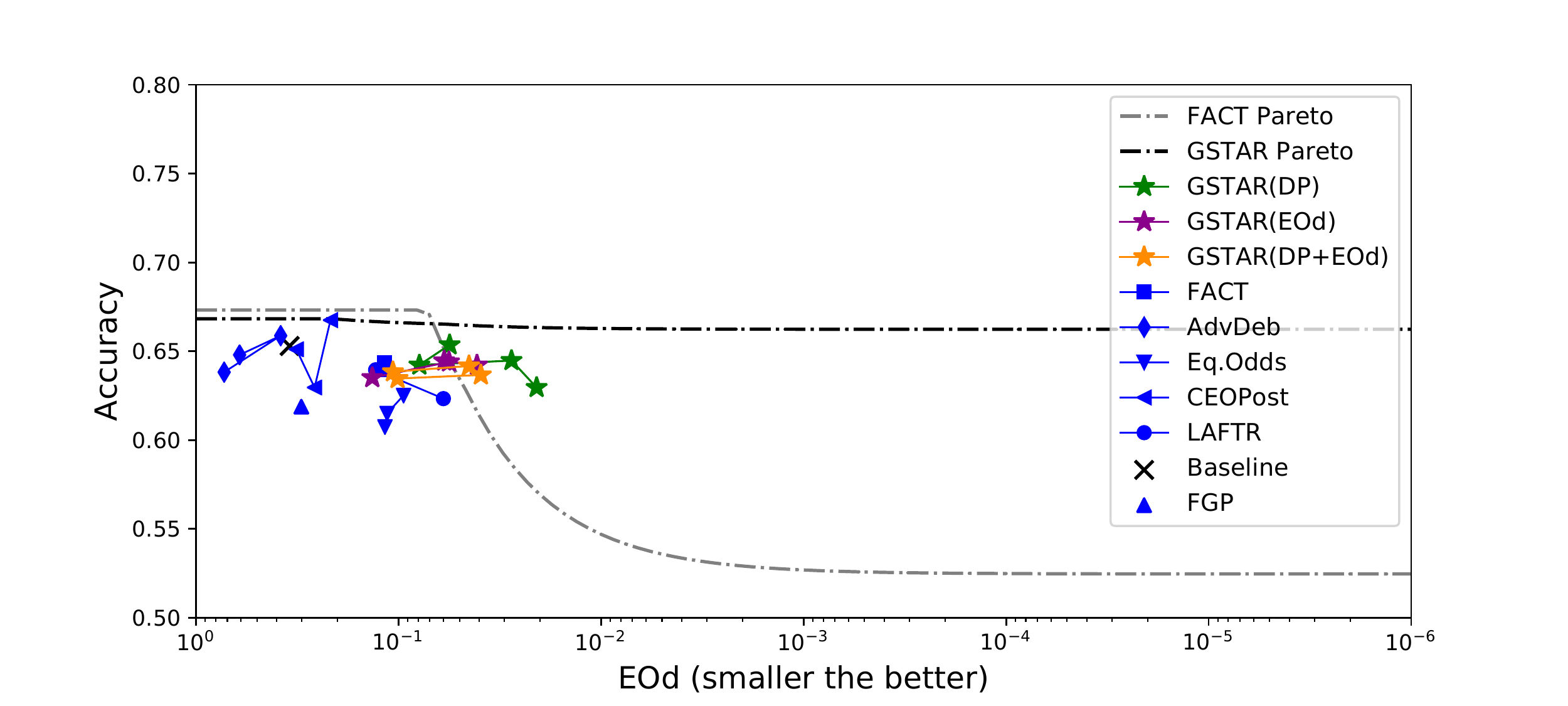}
         \subcaption{Compas Dataset}
         \label{fig:front_compas}
	\end{minipage}
    \begin{minipage}[t]{0.5\textwidth}
		\centering
         \includegraphics[width=1.0\textwidth, height=0.43\textwidth]{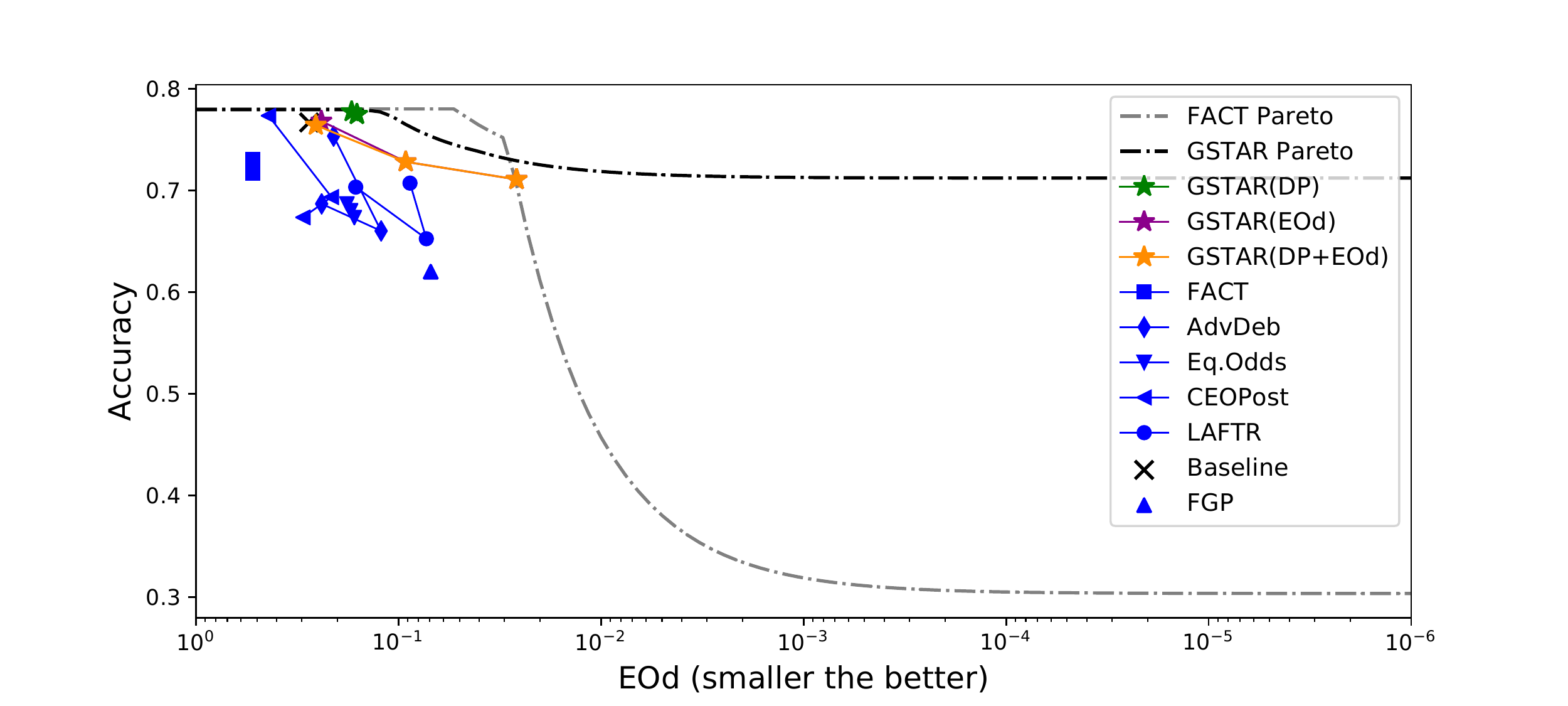}
         \subcaption{German Dataset}
         \label{fig:front_german}
	\end{minipage}	
	\caption{Pareto frontiers of equalized odds to show the upper bound of best achievable accuracy under different fairness constraints. Upper right region under the boundary is desired.
	For three variations of GSTAR with different fairness objectives, GSTAR (stars) is the closest to the Pareto frontier which indicates the best trade-offs.}
    \label{fig:frontier}
\end{figure}

We compare with multiple 
fairness approaches in the experiments. For clear demonstration of results, we use different shapes of marker for each comparing methods in~\autoref{fig:frontier} and~\autoref{fig:post-process}. The comparing methods include:
    FGP (Tan \etal ~\shortcite{tan2020learning}),
    FACT (Kim \etal ~\shortcite{kim2020model}),
    DIR (Feldman \etal  ~\shortcite{feldman2015certifying}),
    AdvDeb (Zhang \etal  ~\shortcite{zhang2018mitigating}),
    CEOPost (Pleiss \etal  ~\shortcite{pleiss2017fairness}),
    Eq.Odds (Hardt \etal  ~\shortcite{hardt2016equality}),
    LAFTR (Madras \etal ~\shortcite{madras2018learning}),
    and Baseline: For CelebA dataset, we use ResNet50~\cite{he2016deep} as a reference, and logistic regression for all other datasets.

We choose broadly used fairness metrics in evaluation including: \textbf{equal opportunity difference} (EOp) and \textbf{equalized odds difference} (EOd) (Hardt \etal ~\shortcite{hardt2016equality}); \textbf{1-disparate impact} (1-DIMP) (Barocas \etal ~\shortcite{barocas2016big}); \textbf{balanced accuracy difference} (BD). 

\begin{figure*}[!t]
     \centering
   \begin{minipage}[t]{0.49\textwidth}
         \centering
         \includegraphics[width=\textwidth, height=0.5\textwidth]{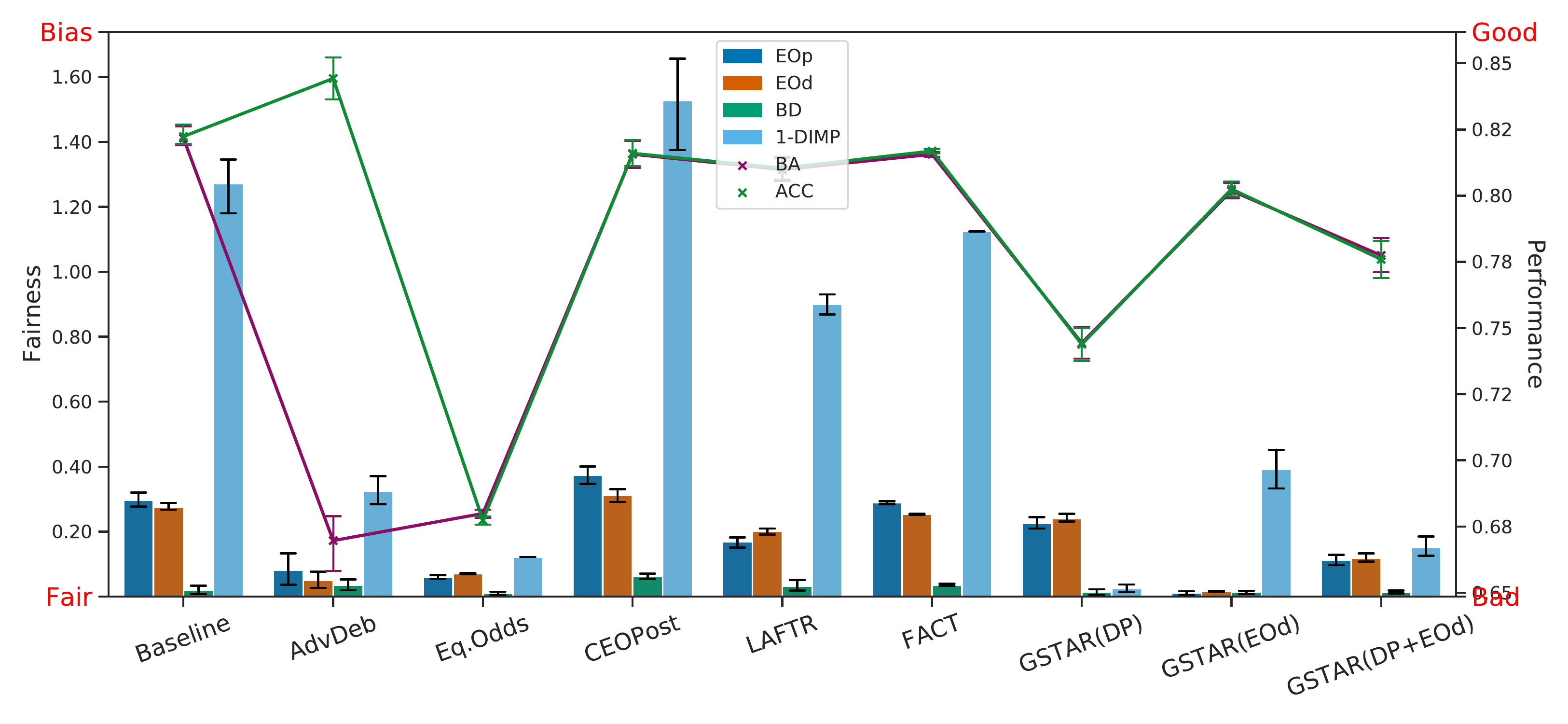}
         \subcaption{CelebA Dataset}
         \label{fig:multi_celeba}
     \end{minipage}
     \begin{minipage}[t]{0.49\textwidth}
         \centering
         \includegraphics[width=\textwidth, height=0.5\textwidth]{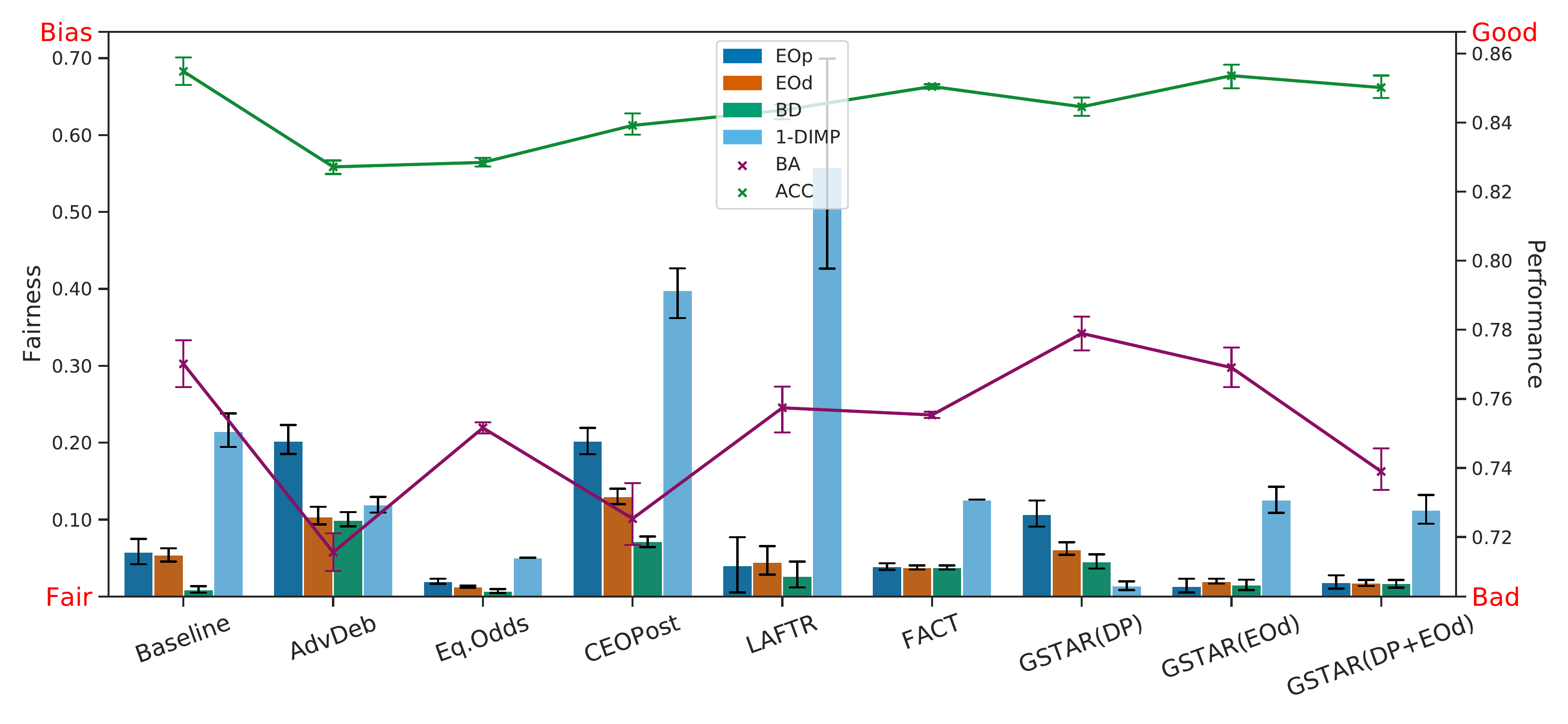}
         \subcaption{Adult Dataset}
         \label{fig:multi_adult}
     \end{minipage}
     \hfill
     \begin{minipage}[t]{0.49\textwidth}
         \centering
         \includegraphics[width=\textwidth, height=0.5\textwidth]{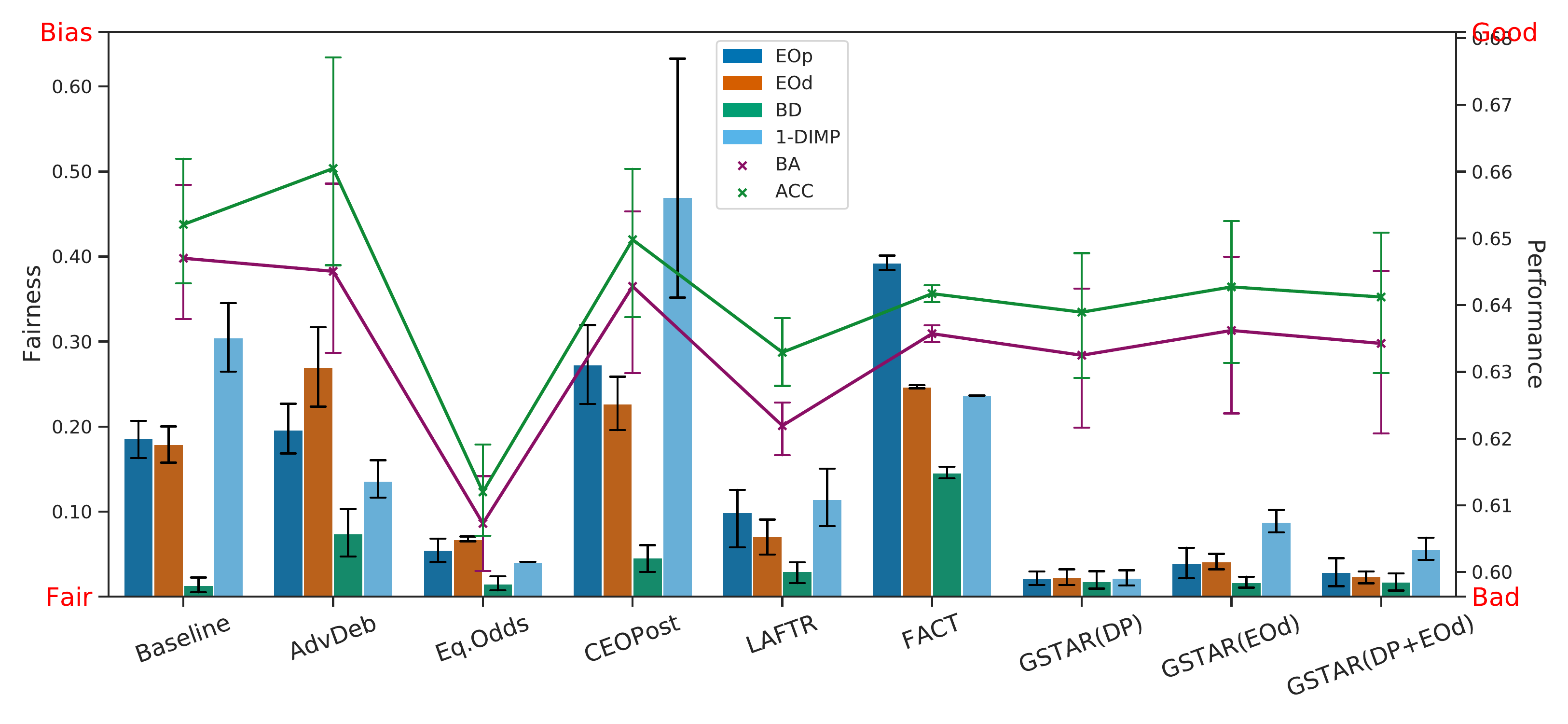}
         \subcaption{Compas Dataset}
         \label{fig:multi_compas}
     \end{minipage}
     \hfill
     \begin{minipage}[t]{0.49\textwidth}
         \centering
         \includegraphics[width=\textwidth, height=0.5\textwidth]{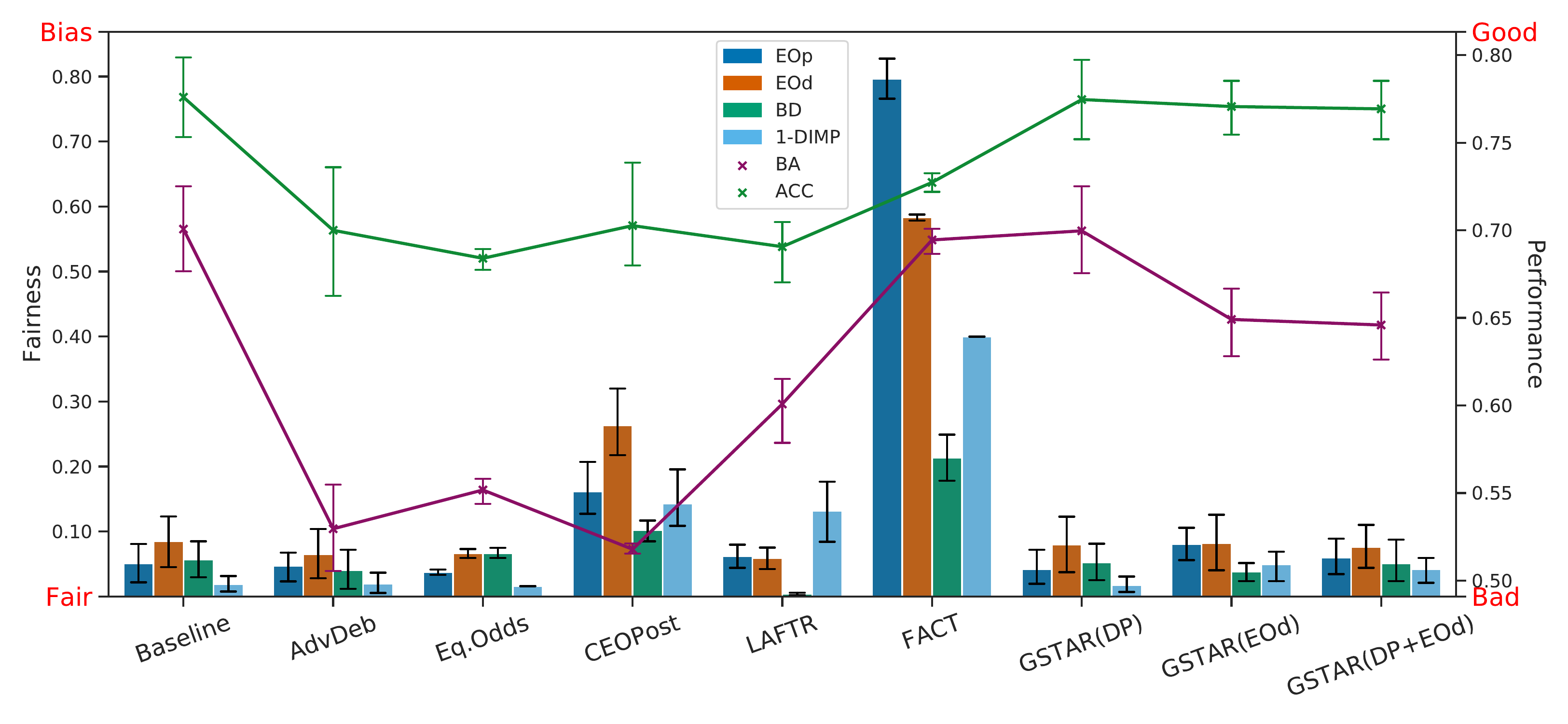}
         \subcaption{German Dataset}
         \label{fig:multi_german}
     \end{minipage}
    \caption{Evaluation on fairness and performance metrics.
    The bar plots indicate fairness measures of each model.
    The line plots indicate the performance measure of each model. 
    Lower fairness values (left y-axis) and higher performance values (right y-axis) show better fairness and performance respectively.
    We consider three variations of GSTAR models (DP, EOd, DP+EOd).}
    \label{fig:multi}
\end{figure*}
\begin{figure*}[!t]
	\centering
	\begin{minipage}[t]{0.49\textwidth}
		\centering
		\includegraphics[width=1\linewidth, height = 0.45\linewidth]{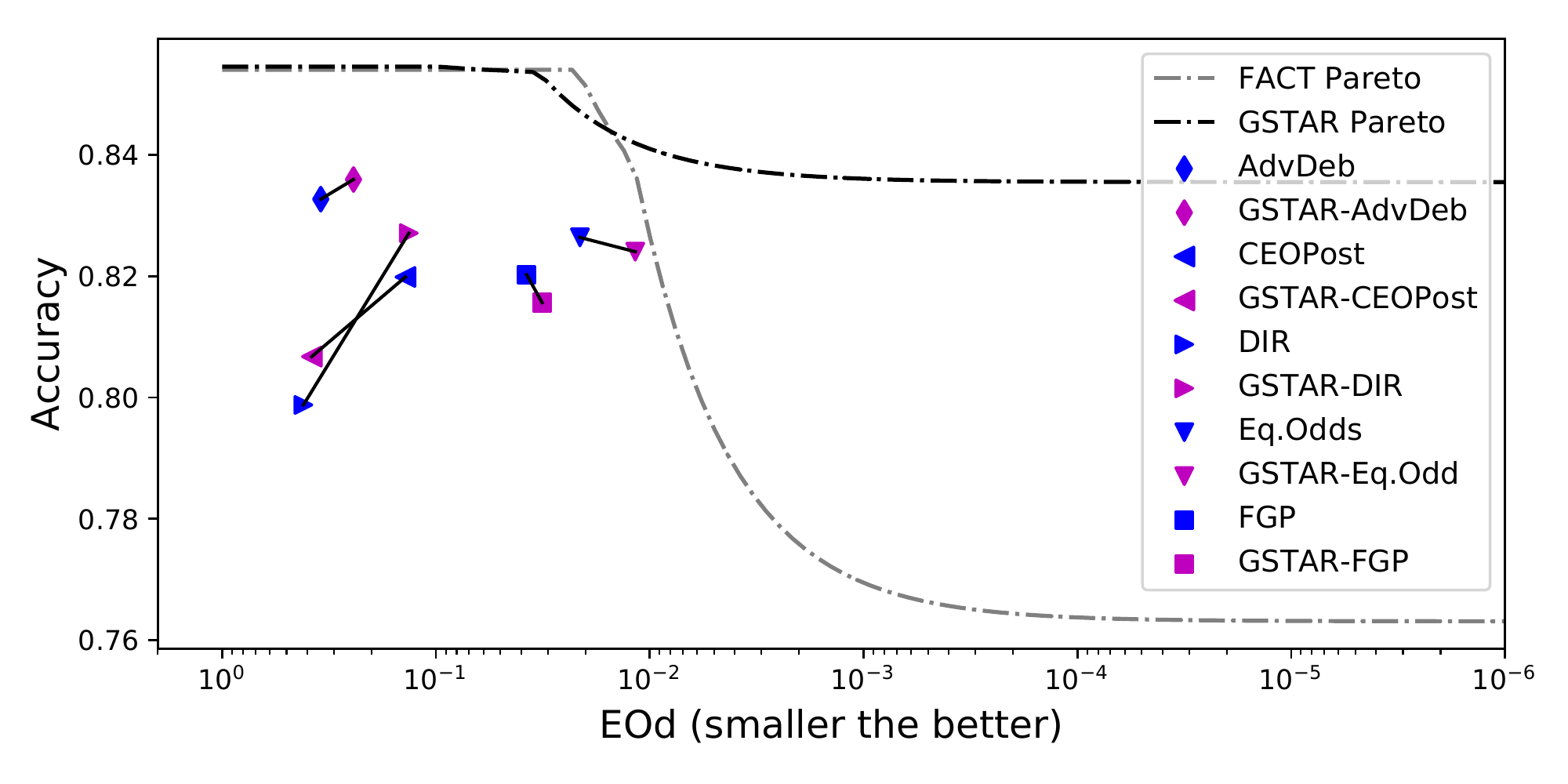}
		\subcaption{Adult Dataset}
	\end{minipage}
    \hfill
	\begin{minipage}[t]{0.49\textwidth}
		\centering
		\includegraphics[width=1\linewidth, height = 0.45\linewidth]{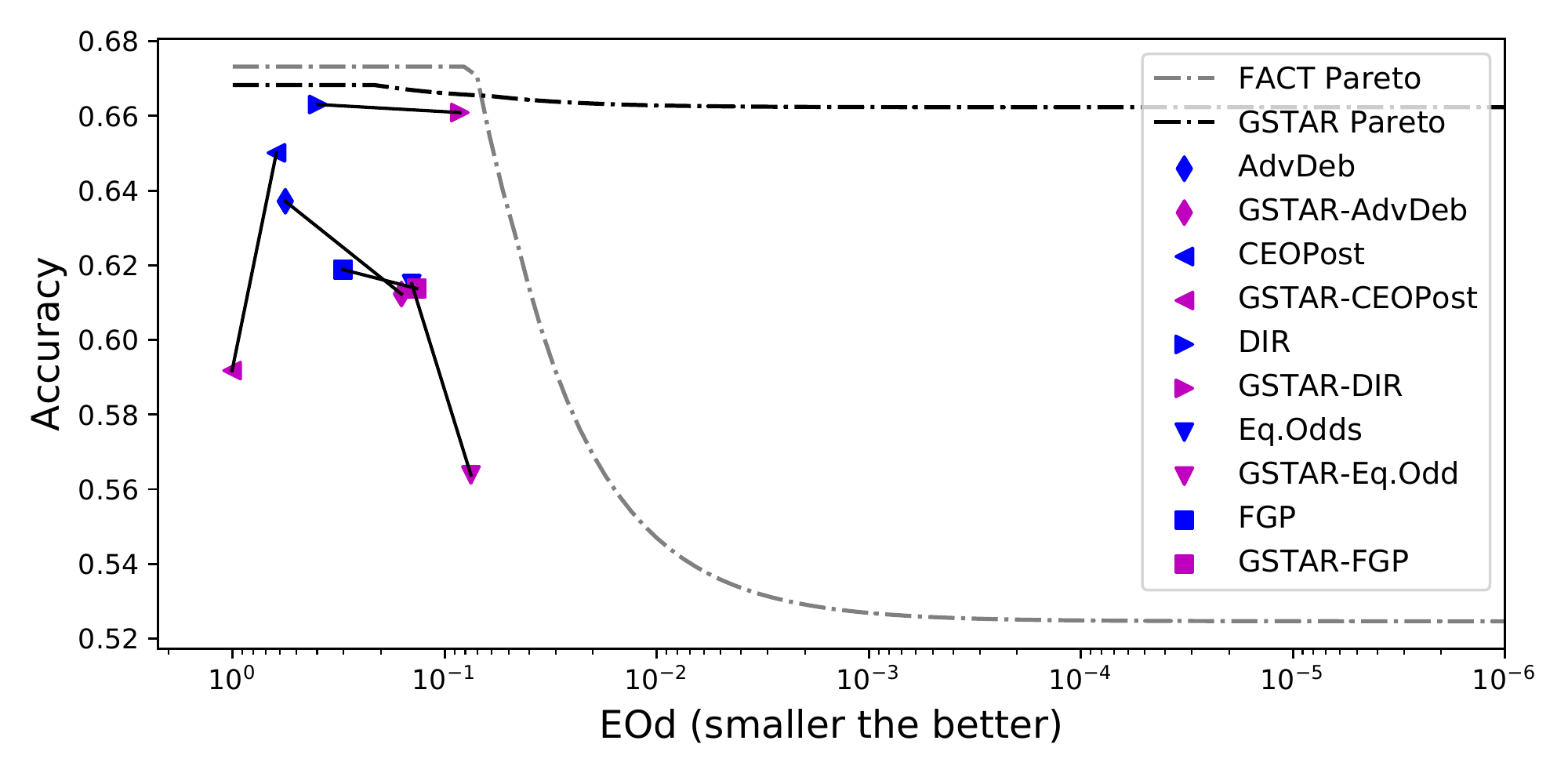}
		\subcaption{Compas Dataset}
	\end{minipage}

	\caption{Illustration of post-processing (magenta colored points) on existing fairness models (blue colored points). Given the outputs of each model, we efficiently improve existing fairness models with optimized group-aware thresholds from GSTAR.}
	\label{fig:post-process}
\end{figure*}

We evaluate the methods on four fairness datasets: 
\textbf{CelebA} dataset ~\cite{celeba},
\textbf{Adult} dataset ~\cite{adult1996},
\textbf{COMPAS}\footnote{\url{https://github.com/propublica/compas-analysis}} dataset, and 
\textbf{German} dataset \cite{Dua2019}.
More details of the comparing methods, evaluation metrics, and datasets are provided in the Supplementary material.






\subsection{Performance and Fairness-Accuracy Trade-Offs} \label{sec:trade-off}
In this subsection, we look into the performance evaluation of GSTAR comparing with other state-of-the-art methods.
We consider Pareto frontier to visualize the trade-offs between fairness and accuracy to demonstrate the measure of performance.

In~\autoref{fig:frontier}, we plot Pareto frontier, which is the upper bound for the accuracy-fairness trade-offs, desired output locates at the upper right region under the boundary which corresponds to higher values in accuracy and lower values in fairness discrepancy.
With the same fairness constraints are given, we achieve a better frontier than the FACT (Kim \etal ~\shortcite{kim2020model}) as we equally weigh on demographic statistics and have a better feasible region.
To obtain our results (star points), we first estimate the logit distribution from the output of the baseline model, and then we get optimal adaptive thresholds with corresponding fairness metric by updating w.r.t. the objective function in~\eqref{eq:obj}.
Here we have three combinations of fairness imposed to GSTAR: demographic parity (DP), equalized odds (EOd), and with both constraints (DP+EOd).
By post-processing on a simple baseline, we achieved significantly better fairness with small or no sacrifice in accuracy.
In all datasets, GSATR got competitive or better results than other state-of-the-art methods on both fairness and accuracy.

For example, {we got $\ttheta_{EOd}^*=(0.640, -0.627)^\mathsf{T}$ for} the CelebA dataset.
This shows that we have a higher threshold for the privileged group and a lower threshold for the unprivileged group.
This optimal thresholding from GSTAR allows more samples from the privileged group to be correctly predicted as unattractive that would compensate for the discrimination of the original model.
In other words,  this improves predictive equality~\cite{chouldechova2017fair} with a huge amount from 0.235 to 0.014. 
Also, true positive rate difference (also known as equality of opportunity~\cite{hardt2016equality}) got reduced from 0.282 to 0.018.
It is notable that GSTAR only sacrificed $2.2\%$ of accuracy to bring the big improvement in fairness.


Since the objective function of our model is independent to data dimensionality, our model is much more efficient especially for high dimensional data.
We mostly outperform the computational cost comparing to the other methods.
The comparison of computational time and auxiliary experiments on the datasets can be found in the Supplementary material.

\subsection{Flexibility and Multiple Fairness Constraints}
\label{sec:bar_plot}
Since each fairness metric has different interests, it has been theoretically proven that they cannot be perfectly satisfied all together~\cite{pleiss2017fairness, chouldechova2017fair, kleinberg2016inherent}.
Because of this inherent trade-offs between fairness metrics, most of the recent works focus on a single metric at a time to achieve fairness.
However with GSTAR, we have the flexibility to optimize on multiple fairness constraints that can be represented in the confusion matrix format.
Moreover, given the estimated distribution $f_{ya}$ of a arbitrary classification model, we can adjust the optimal $\ttheta$ based on the needs by accommodating different fairness criteria.

\autoref{fig:multi} demonstrates the result of the methods with fairness metrics and accuracy trade-off evaluations.
Overall, the variations of GSTAR achieve the best fairness on each target fairness while preserving the performance.
For example in \autoref{fig:multi_celeba}, GSTAR with EOd constraint has outstanding performance in most fairness metrics with comparable accuracy ($80.3\%$). Comparing with GSTAR (EOd), when we introduce EOd and DP together (DP+EOd), we achieve significantly better w.r.t. DP fairness with sacrificing a small amount of accuracy and EOd. 

In general, by sacrificing individual fairness performance, we could introduce multiple constraints.
Also, we observe that the more fairness constraints are introduced, the more accuracy is sacrificed.
We empirically found that in some cases (\eg ~\autoref{fig:multi_compas}), introducing multiple fairness is complementary to each other that improves both conditions.

\subsection{Post-Processing on an Existing Fair Model}
For a binary classifier that has a single fixed classification threshold ($0$ for out logit, and $0.5$ for label probability), we can provide better trade-off between fairness and accuracy with GSTAR. Given the logit/probability in the model-agnostic manner, we can improve the fairness as illustrated in~\autoref{fig:post-process}. In most cases, we observe improvement in fairness after GSTAR post-processing.
It is also interesting to note that by optimizing the different thresholds for each protected group, we even obtain better performance on both fairness and accuracy, which indicates that the threshold optimization can not only improve fairness but also accuracy.

However, when the distribution of the logits/probability is highly extreme (such as the results of using GSTAR to post-process CEOPost), it is difficult to estimate the distribution and thus causes erroneous optimization in GSTAR. We empirically found that when the dataset is extremely imbalanced such that we do not have enough samples to estimate the logit/probability distribution, or the given classification model is too certain to the prediction that samples are concentrated to certain output, this problem arises.

\section{Conclusion and Discussion}
\label{sec:conclusion}
In this paper, we propose a group-aware threshold adaptation method (GSTAR) to post-process in model-agnostic manner and optimize over multiple fairness constraints.
We directly optimize the classification threshold for each demographic group w.r.t. the classification error and multiple fairness constraints in a unified objective function, such that we can practically achieve an optimal trade-off between accuracy and fairness in fair classification.
Our method is applicable to diverse notions of group fairness as the majority of fairness notions can be expressed as a linear or quadratic equation through confusion matrix.
We empirically show that GSTAR is \textit{flexible} with fairness regularization, \textit{efficient} with low computational cost.
We also notice that the adaptive thresholds benefit accuracy in some cases. 
GSTAR agrees to protect \textit{privacy} such as {article 17 of EU's GDPR~\cite{regulation2016regulation}}. We only require the estimated distribution of the output from a given model i.e., our post-processing method is oblivious to features. Thus training data is no longer needed and allowed to be discarded after training the model that to be post-processed. Thus, GSTAR can be applied to relaxed scenarios where practitioners cannot access individual-level sensitive information but have estimated distributions of logits for each sensitive group. 

Further, we empirically find that GSTAR is not applicable to post-process some classification models in the following situations:
1) the model does not provide logit/probability as the outcome;
2) The model provides an extreme distribution of the output logit/probability. For example, when the model is too certain about its prediction, it will be difficult to perform probability density estimation.
In our future work, we will study possible strategies to solve the above limitations, and extend GSTAR to multi-class, multi-sensitive group problems and improve the fairness-accuracy trade-off in a more general scheme.
\bibliographystyle{aaai22}
\bibliography{egbib}

\twocolumn[{
\begin{center}
  \centering
  \vskip 60pt
  \LARGE Supplementary Material for ``Group-Aware Threshold Adaptation for Fair Classification'' \par
  \vskip 4em
\end{center}}]

\setcounter{figure}{4}
\setcounter{secnumdepth}{2}
\renewcommand\thesection{\arabic{section}}
\setcounter{section}{5}

\section{Optimization Procedure of GSTAR} \label{sec:optim_method}
The threshold $\boldsymbol{\theta}$ is optimized with alternating optimization method in GSTAR. 
Here we take EOp constraint as an example to show the alternating optimization steps, then $\LL_{fair}(\ttheta)$ can be written as
\begin{equation}
    \LL_{fair}^{EOp}(\ttheta) = \left(\tp_1(\theta_1) - \tp_0(\theta_0)\right)^2,
\end{equation}
and overall objective is to minimize
\begin{equation}
\label{eq-sup:obj}
\LL(\ttheta) = \LL_{per}(\ttheta) + \lambda \LL_{fair}^{EOp}(\ttheta).
\end{equation}
\textbf{The first step} is to fix $\theta_0$ and update $\theta_1$. We can approximate the terms that are related to $\theta_1$ (\eg $\tp_1, \fp_1, \tn_1, \fn_1$) {in~\eqref{eq-sup:the-con}} with first-order Taylor expansion at $\theta_1^{\tau-1}$.
For example, 
\begin{eqnarray}
\label{eq-sup:taylor}
    \tp_1(\theta_1) \approx \tp_1(\theta_1^{\tau-1}) + \frac{\partial \tp_1}{\partial  \theta_1}\Big|_{\theta_1 = \theta_1^{\tau-1}}
    (\theta_1-\theta_1^{\tau-1})
\end{eqnarray}
From~\eqref{eq-sup:the-con}, we can easily derive that
\begin{eqnarray}
\begin{aligned}
\tp_1(\theta_1^{\tau-1}) &=&& 1 - \int_{-\infty}^{\theta_1^{\tau-1}}f_{11}(x)dx,
\\
\frac{\partial \tp_1}{\partial \theta_1} &=&& -f_{11}(\theta_1^{\tau-1}).
\end{aligned}
\end{eqnarray}

Similarly, we can find the first order Taylor expansion of $\fp_1, \fn_1$, \text{and} $\tn_1$.
Then, the update of $\theta_1$ w.r.t.~\eqref{eq-sup:obj} can be approximated with the following minimization problem w.r.t. $\Delta_1$
\begin{eqnarray}
\label{eq-sup:obj_delta1}
\Delta^\tau_1 := \argmin_{\Delta_1} (\eta_1^\tau + \alpha_1^\tau \Delta_1)^2 + \lambda(\epsilon_1^\tau + \beta_1^\tau \Delta_1)^2, 
\end{eqnarray}
where $\Delta_1 = \theta_1 - \theta_1^{\tau-1}$ and 
\begin{eqnarray}
\label{eq-sup:update}
\begin{aligned}
\alpha_1^\tau =& \frac{n_{11}}{N} f_{11}(\theta_1^{\tau-1}) -  \frac{n_{01}}{N} f_{01}(\theta_1^{\tau-1}),    \beta_1^\tau = -f_{11}(\theta_1^{\tau-1}),
\\
\eta_1^\tau =& \int_{-\infty}^{\theta_1^{\tau-1}} \big( \frac{n_{11}}{N}f_{11}(x) + \frac{n_{01}}{N} (1-f_{01}(x)
\big)dx \\
&  +\int_{-\infty}^{\theta_0^{\tau-1}} \big( \frac{n_{10}}{N}f_{10}(x) + \frac{n_{00}}{N} (1-f_{00}(x) \big)dx,
\\
\epsilon_1^\tau =& \int_\infty^{\theta_1^{\tau-1}}f_{11}(x)dx - \int_\infty^{\theta_0^{\tau-1}}f_{10}(x)dx .
\end{aligned}
\end{eqnarray}

Taking the derivative of~\eqref{eq-sup:obj_delta1} w.r.t. $\Delta_1$ and setting it to 0, we can easily obtain the closed-form solution of $\Delta_1^\tau$ as
\begin{equation}
\Delta_1^\tau = -\frac{\alpha^\tau\eta^\tau + \lambda\beta^\tau\epsilon^\tau}{{ (\alpha^\tau)^2 + \lambda(\beta^\tau)^2}}.
\label{eq-sup:update_delta}
\end{equation}
\textbf{The second step} is to fix $\theta_1$ and update $\theta_0$, and this can be achieved in a similar way of updating $\theta_1$. Then we can finalize the alternating optimization as:
\begin{equation}
\label{eq-sup:opt}
\begin{split}
    &\theta_0^{\tau} = \theta_0^{\tau-1} + \Delta_0^{\tau}, \qquad \theta_1^{\tau} = \theta_1^{\tau-1} + \Delta_1^{\tau}.
\end{split}
\end{equation}
It is notable that in each iteration we derive the optimal update step $\Delta_a$, which eliminates the burden of tuning hyperparameter (such as learning rate) in iterative algorithm.
The optimization step is summarized in Algorithm~\ref{alg}. The above algorithm can easily extend to multiple fairness constraints by adding corresponding squared-loss fairness terms to~\eqref{eq-sup:obj}. 

\begin{algorithm}[!t]
\caption{Optimization Algorithm of GSTAR Model}
\begin{algorithmic} \label{alg}
	\STATE \textbf{Input} dataset $\X\times\A\times\Y = \{(\x_i,\a_i, \y_i)\}_{i=1}^n$, classification model $h(X)$, hyperparameter $\lambda$.
	\STATE \textbf{Output} Group-specific threshold $\ttheta = (\theta_1, \theta_0)$.
	\STATE \textbf{Initialize} $\ttheta = (\theta_1, \theta_0) = (0,0)$.
	
	\STATE 1. Given a classifier $H(x)$, estimate probability density function $f_{ya}, y,a \in \{0,1\}$ 
	by maximum likelihood estimation.

	\WHILE{not converge}{
    
    \STATE 2. Calculate the optimal step $\Delta_1$ as
    $\Delta_1 =  -\frac{\alpha_1\eta_1+\lambda\beta_1\epsilon_1}{\alpha_1^2 + \lambda \beta_1^2}$, with $\alpha_1, \beta_1, \eta_1, \epsilon_1$ values shown in~\eqref{eq-sup:update};
    
    \STATE 3. {Update the threshold:} 
    $\theta_1 \leftarrow \theta_1 + \Delta_1$;
    
    
    \STATE 4. Calculate the optimal step $\Delta_0$ as $\Delta_0 =  -\frac{\alpha_0\eta_0+\lambda\beta_0\epsilon_0}{\alpha_0^2 + \lambda \beta_0^2}$ with $\alpha_0, \beta_0, \eta_0, \epsilon_0$ values calculated in a similar way as in~\eqref{eq-sup:update}:
    
    \STATE 5.  {Update the threshold:}
    $\theta_0 \leftarrow \theta_0 + \Delta_0$.
	}\ENDWHILE
\end{algorithmic}
\end{algorithm}

\section{Upper Bounds on False-Positive/Negative Rate Gap Between Groups}
\label{app:proof-theorem1and2}

\subsection{Notations}
We start from defining notations. 
We denote $f_{ya}(x)$ for the estimated parametric probability density function (PDF) of the distribution of output logit $h$ in the subset $\{Y = y, A = a\}$.
Correspondingly, we denote the corresponding cumulative distribution function (CDF) as
\[
F_{ya}(x) = \int_{-\infty}^x f_{ya}(x) dx. 
\]
We use $F^{-1}_{ya}(x)$ to denote the inverse of the CDF. 

Then, following the definitions given in the main paper, we have 
\begin{eqnarray}
\begin{aligned}
\label{eq-sup:the-con}
    &\text{TP}_a(\theta_a) = 1- F_{1a}(\theta_a), \qquad
    &\text{FN}_a(\theta_a) = F_{1a}(\theta_a), \\
    &\text{FP}_a(\theta_a) = 1- F_{0a}(\theta_a), \qquad
    &\text{TN}_a(\theta_a) = F_{0a}(\theta_a) . 
\end{aligned}
\end{eqnarray}

\subsection{Characterizing the accuracy loss function under perfect EOp condition}
\label{app:charact-eop}

Before stating the theorem, we illustrate the difference between $\LL_{per}(\ttheta)$ used in our paper versus loss function one would use in a population-wise classification problem (without considering group-aware thresholds). That is, one would only consider the loss function on accuracy
\begin{equation}
\bar{\LL}_{per}(\theta)
= \left( r_1 \bar{\fn}(\theta) + r_0 \bar{\fp}(\theta) \right)^2,
\label{eq-sup:accu-loss-pop}
\end{equation}
where only one threshold $\theta$ (for both groups) needs to be decided, $r_y = (n_{y0}+n_{y1})/N$ is the population ratio of data samples with label $y$, $\bar{\fn}(\theta), \bar{\fp}(\theta)$ are the population-wise false-negative and false-positive rate. $\bar{\fn}(\theta), \bar{\fp}(\theta)$ are defined in a similar way as in~\eqref{eq-sup:the-con} except that we just use the population-wise pdf $\bar{f}_y(x)$ in the integral for label $y$. \eqref{eq-sup:accu-loss-pop} will be our benchmark to compare with $\LL_{per}(\ttheta)$ used in our paper.

We start from considering the case that we achieve perfect EOp condition, that is 
\begin{equation}
\label{eq-sup:tp-perfect}
\tp_1 (\theta_1) = \tp_0 (\theta_0),
\end{equation}
or equivalently 
\begin{equation*}
    \text{FN}_1 (\theta_1) = \text{FN}_0 (\theta_0).
\end{equation*}
This means that $\theta_0$ and $\theta_1$ satisfies the following condition
\begin{eqnarray}
\label{eq-sup:tp-perfect-theta}
F_{11}(\theta_1) = F_{10}(\theta_0).
\end{eqnarray}
Equivalently, we have 
\begin{eqnarray}
\label{eq-sup:tp-perfect-theta-equiv}
\theta_0 = F^{-1}_{10} \big( F_{11}(\theta_1) \big) . 
\end{eqnarray}

Under any given pair of $(\theta_0, \theta_1)$ that satisfies~\eqref{eq-sup:tp-perfect-theta-equiv}, recall that the performance error $\LL_{per}(\ttheta)$ is defined as
\begin{multline}
 \LL_{per}(\ttheta) = \Big(\frac{n_{01}}{N}\fp_1(\theta_1) + \frac{n_{11}}{N}\fn_1(\theta_1) + \\\frac{n_{00}}{N}\fp_0(\theta_0) + \frac{n_{10}}{N}\fn_0(\theta_0)\Big)^2 . 
\label{eq-sup:recall-acc-loss}
\end{multline}
From~\eqref{eq-sup:tp-perfect}, we get
\begin{eqnarray*}
\frac{n_{11}}{N}\fn_1(\theta_1) + \frac{n_{10}}{N}\fn_0(\theta_0)
=& \frac{n_{11}+n_{10}}{N} {\fn}(\theta_1) \\
=& r_1 {\fn}(\theta_1), 
\end{eqnarray*}
where $r_1$ denotes, over the entire population (across different groups), proportion of samples with positive labels. In other words, $r_1 {\fn}(\theta_1)$ represents the proportion of data samples (from both groups) with positive label but falsely classified as negative out of the entire dataset.

Next, we look at the other two terms:
\begin{equation*}
\frac{n_{01}}{N}\fp_1(\theta_1)
+ \frac{n_{00}}{N}\fp_0(\theta_0) . 
\end{equation*}
This sum can be written as
\begin{eqnarray*}
& \frac{n_{01}}{N}\fp_1(\theta_1)
+ \frac{n_{00}}{N}\fp_0(\theta_0)
\\
=&  \frac{n_{01}+n_{00}}{N}\fp_1(\theta_1)
+ \frac{n_{00}}{N}\big( \fp_0(\theta_0)
- \fp_1(\theta_1) \big) 
\\
=&  r_0 {\fp}(\theta_1) 
+ \frac{n_{00}}{N}\big( \fp_0(\theta_0)
- \fp_1(\theta_1) \big) . 
\end{eqnarray*}
We denote $\epsilon_1 = \big( \fp_0(\theta_0)
- \fp_1(\theta_1) \big)$. Hence, 
\begin{eqnarray}
\begin{aligned}
 \LL_{per}(\ttheta) = \LL_{per}(\theta_1) = 
\left( r_1 {\fn}(\theta_1) +
 r_0 {\fp}(\theta_1)
 + \frac{n_{00}}{N}\epsilon_1  
\right)^2 . 
\end{aligned}
\label{eq-sup:accu-loss-constraint}
\end{eqnarray}

Comparing~\eqref{eq-sup:accu-loss-pop} with~\eqref{eq-sup:accu-loss-constraint}, we can see that, when $\fp_0(\theta_0)
> \fp_1(\theta_1)$, the term $\frac{n_{00}}{N}\epsilon_1$ captures the additional accuracy loss due to that we have chosen two different thresholds even though that condition~\eqref{eq-sup:tp-perfect-theta} is satisfied. Next, we characterize an upper bound for $\epsilon_1$.

\subsection{Theorem 1 and its Proof}



\begin{proof} 
Recall that $\fp_1(\theta_1) = 1 - F_{01}(\theta_1)$ and $\fp_0(\theta_0) = 1 - F_{00}(\theta_0)$. Hence, 
\begin{eqnarray*}
\big| \fp_0(\theta_0)
- \fp_1(\theta_1) \big| &=&
\big| F_{01}(\theta_1) - F_{00}(\theta_0) \big| \\ 
&\leq& \big| F_{01}(\theta_1) - F_{01}(\theta_0) \big| \\
& & + \big| F_{01}(\theta_0) - F_{00}(\theta_0) \big| .
\end{eqnarray*}
To bound $\epsilon$, we just need to bound $\big|F_{01}(\theta_1) - F_{01}(\theta_0) \big|$ and $\big|F_{01}(\theta_0) - F_{00}(\theta_0) \big|$. 

For the second one, we note that from Assumption 1 that 
$$\big|F_{01}(\theta_0) - F_{00}(\theta_0) \big|
\leq u_0. $$

For the first one, we note that 
\begin{eqnarray*}
\big| F_{01}(\theta_1) - F_{01}(\theta_0)\big|
&\leq& \hat{f}_{01} |\theta_1 -\theta_0|,
\end{eqnarray*}
where $\hat{f}_{01}=\max_x f_{01}(x)$. 

Next, we bound $|\theta_1 -\theta_0|$. 
Note that from~\eqref{eq-sup:tp-perfect-theta-equiv}, 
\begin{eqnarray*}
|\theta_1 -\theta_0| 
&=& \big| F^{-1}_{10} \big( F_{11}(\theta_1) \big) -\theta_1 \big|
\\
&=& \Big| F^{-1}_{10} \big( F_{11}(\theta_1) \big) - F^{-1}_{10}\big( F_{10}(\theta_1) \big) \Big|
\\
&\leq& M_{10} \big| F_{11}(\theta_1) - F_{10}(\theta_1) |
\\
&\leq& M_{10} u_1. 
\end{eqnarray*}
\end{proof}

Theorem 1 provides an upper bound on the difference in the false positive rate between the two groups, for any given pair of $(\theta_0, \theta_1)$ such that the false negative rates are the same for the two groups (\ie satisfies the perfect EOp condition). 
As discussed in Section~\ref{app:charact-eop}, this upper bound also characterize the additional accuracy loss due to that we have group-dependent thresholds compared to the case with only one threshold for both groups.

\subsection{Theorem 2 and its Proof}
For predictive equality (PE) condition, we prove a similar result. That is, assuming we achieve perfect PE condition with 
\begin{equation}
\label{eq-sup:pe_perfect}
\fp_1 (\theta_1) = \fp_0 (\theta_0), 
\end{equation} 
or equivalently
\begin{equation}
\text{TN}_1 (\theta_1) = \text{TN}_0 (\theta_0) .
\end{equation} 
This means that $\theta_0$ and $\theta_1$ satisfies the following condition 
\begin{eqnarray}
\label{eq-sup:pe-perfect-theta}
F_{01}(\theta_1) = F_{00}(\theta_0).
\end{eqnarray}
Equivalently, we have 
\begin{eqnarray}
\label{eq-sup:pe-perfect-theta-equiv}
\theta_0 = F^{-1}_{00} \big( F_{01}(\theta_1) \big) . 
\end{eqnarray}

Under any given pair of $(\theta_0, \theta_1)$ that satisfies~\eqref{eq-sup:pe-perfect-theta-equiv}, the performance error $\LL_{per}(\ttheta)$ can be written as
\begin{eqnarray*}
 \LL_{per}(\ttheta) &=& \Big(\frac{n_{01}}{N}\fp_1(\theta_1) + \frac{n_{11}}{N}\fn_1(\theta_1) \\
 &\qquad &  \qquad \qquad + \frac{n_{00}}{N}\fp_0(\theta_0) + \frac{n_{10}}{N}\fn_0(\theta_0)\Big)^2 
 \\
&=& \left(r_1{\fn}(\theta_1) + r_0{\fp}(\theta_1) 
+ \frac{n_{10}}{N}\epsilon_2 \right)^2, 
\end{eqnarray*}
where 
\[
\epsilon_2 = \big( \fn_0(\theta_0) - \fn_1(\theta_1) \big). 
\]

Similar to Theorem 1, we can provide an upper bound on $\epsilon_2$ under Assumption 1. 

\begin{proof}
The proof is similar to that of Theorem 1. We provide the main steps and omit details that repeat with the proof of Theorem 1. We have 
\begin{eqnarray*}
\big| \fn_0(\theta_0) - \fn_1(\theta_1) \big|
&=& \big| F_{11}(\theta_1) - F_{10}(\theta_0) \big|
\\
&\leq& \big| F_{11}(\theta_1) - F_{11}(\theta_0) \big| \\
&& \quad +\big| F_{11}(\theta_0) - F_{10}(\theta_0) \big|
\\
&\leq& \hat{f}_{11}|\theta_1 -\theta_0|
+ u_1 
\\
&\leq& \hat{f}_{11} M_{00} u_0 + u_1. 
\end{eqnarray*}
\end{proof}
Theorem 2 provides an upper bound on the difference in the false negative rate between the two groups, for any given pair of $(\theta_0, \theta_1)$ such that the false positive rates are the same for the two groups (\ie satisfies the perfect PE condition).

To sum up, Theorem 1 and 2 characterize the upper bound of false positive/negative rate gap between two groups when the false negative/positive rate gap is 0. At the same time, it captures the upper bound of additional accuracy loss due to the two different thresholds for different groups under a perfect fairness (EOp or EP) condition.


\section{Characterizing the trade-off between Accuracy and Fairness}

In this section, we prove a theorem to characterize the trade-off between accuracy and fairness. That is, we start from the perfect EOp or PE conditions and perturb the solution by a small amount. We then bound the difference in the accuracy loss by comparing the perturbed solution with the original solution that satisfies the perfect fairness conditions.

\subsection{Perturbed EOp condition}

To start with, let us consider solutions $(\theta_0, \theta_1)$ that satisfy the perfect EOp condition~\eqref{eq-sup:tp-perfect-theta-equiv}. Under this condition, the optimization problem becomes one dimensional, that is, 
\[
\theta_1^* = \argmin_{\theta_1}  \LL_{per}(\theta_1),  
\]
where 
$$
\LL_{per}(\theta_1) = \left( r_1 \fn_1(\theta_1) + r_0 \fp_1(\theta_1) 
+ \frac{n_{00}}{N} \epsilon_1 (\theta_1) \right)^2 
$$
and 
\begin{eqnarray*}
\epsilon_1(\theta_1) &=& \fp_0(\theta_0) - \fp_1(\theta_1) \\
                     &=& F_{01}(\theta_1) - F_{00}\big(F_{10}^{-1}(F_{11}(\theta_1)) \big) . 
\end{eqnarray*}

From $\theta_1^*$, we can get the corresponding $\theta_0^* = F_{10}^{-1}(F_{11}(\theta_1^*))$. 
We further denote this optimal accuracy loss value as 
\[
L^* = \LL_{per}(\theta_1^*). 
\]

Now with the optimal solution $(\theta_0^*,\theta_1^*)$, we investigate the changes in $\LL_{per}(\theta_1^*)$ when we perturb the perfect EOp condition and allow a small difference. That is, now consider solution $(\tilde{\theta}_0, \tilde{\theta}_1)$ such that 
\begin{equation}
\begin{split}
  |\fn_1({\theta_1}^*) - \fn_1({\tilde{\theta}_1})|
\leq \gamma/2, \\
|\fn_0({\theta_0}^*) - \fn_0({\tilde{\theta}_0})|
\leq \gamma/2.    
\end{split}
\label{eq-sup:perturb-assump}
\end{equation}
Consequently, the solution $(\tilde{\theta}_0, \tilde{\theta}_1)$ satisfy the following perturbed EOp condition:
\begin{equation}
\big| \tp_1(\tilde{\theta}_1) - \tp_0(\tilde{\theta}_0) \big|
= \big| \fn_1(\tilde{\theta}_1) - \fn_0(\tilde{\theta}_0) \big|
\leq \gamma. 
\end{equation}
Without loss of generality, we assume that (i) the true positive rate of group 1 is higher than that of group 0, and (ii) the above inequality is binding (because if not binding, then we can always choose a smaller $\gamma$ to make it binding). Thus, we have $\tp_1(\tilde{\theta}_1) - \tp_0(\tilde{\theta}_0)=\gamma$, or equivalently, $\fn_0(\tilde{\theta}_0)- \fn_1(\tilde{\theta}_1) = \gamma$. 
This gives us 
\begin{equation}
\tilde{\theta}_0 = F_{10}^{-1}
\big(F_{11}(\tilde{\theta}_1)+\gamma) \big).  
\end{equation}

Next, we analyze $\LL_{per}(\tilde{\theta}_1)$ by substituting $(\tilde{\theta}_0, \tilde{\theta}_1)$ in~\eqref{eq-sup:recall-acc-loss}, which gives us 
$$
\LL_{per}(\tilde{\theta}_1) = \left( r_1 \fn_1(\tilde{\theta}_1) + r_0 \fp_1(\tilde{\theta}_1) 
+ \frac{n_{10}}{N} \gamma 
+ \frac{n_{00}}{N} \tilde{\epsilon}_1 (\tilde{\theta}_1) \right)^2 , 
$$
where 
\begin{equation*}
\begin{split}
\tilde{\epsilon}_1(\tilde{\theta}_1) &= \fp_0(\tilde{\theta}_0) - \fp_1(\tilde{\theta}_1) \\
&= F_{01}(\tilde{\theta}_1) - F_{00}\big(F_{10}^{-1} \big(F_{11}(\tilde{\theta}_1)+\gamma) \big) \big) . 
\end{split}
\end{equation*}

We denote the optimal value for this perturbed version as $\tilde{\theta}_1^*$, and its corresponding loss value as 
\[
\tilde{L}^* = \LL_{per}(\tilde{\theta}_1^*).
\]
Furthermore, from~\eqref{eq-sup:perturb-assump}, we have 
\begin{equation}
|\fn_1({\theta_1^*}) - \fn_1({\tilde{\theta}_1^*})|
=|F_{11}(\theta_1^*) - F_{11}({\tilde{\theta}_1^*}) | \leq \gamma/2.  
\end{equation}
Under Assumption 1, we have 
\begin{eqnarray*}
|\theta_1^* - \tilde{\theta}_1^*|
&=& \big| F_{11}^{-1}(F_{11}(\theta_1^*)) - F_{11}^{-1}(F_{11}({\tilde{\theta}_1^*})) \big|
\\
&\leq& M_{11}\big| F_{11}(\theta_1^*) - F_{11}({\tilde{\theta}_1^*}) \big|
\\
&=& M_{11} \gamma/2. 
\end{eqnarray*}

\subsection{Theorem 3 and its proof}

We are ready to compare $\LL_{per}(\theta_1^*)$ and $\LL_{per}(\tilde{\theta}_1^*)$. The latter loss should be no larger than the former since we relaxed the perfect EOp condition (constraint) in the optimization, \ie $L^*\geq\tilde{L}^*$.

\begin{proof}
We have that
\begin{eqnarray*}
&&\LL_{per}(\theta_1^*) - \LL_{per}(\tilde{\theta}_1^*)
\\
&\leq& 2 L^*   
\Big| r_1 \fn_1(\theta_1^*) + r_0 \fp_1(\theta_1^*) 
+ \frac{n_{00}}{N} \epsilon_1 (\theta_1^*)
\\
&&- \Big( r_1 \fn_1(\tilde{\theta}_1^*) + r_0 \fp_1(\tilde{\theta}_1^*) 
+ \frac{n_{10}}{N} \gamma 
+ \frac{n_{00}}{N} \tilde{\epsilon}_1 (\tilde{\theta}_1^*) \Big) 
\Big| 
\\
&\leq& 2L^* \big( r_1 \gamma/2
+ r_0|\fp_1(\theta_1^*)-\fp_1(\tilde{\theta}_1^*)|\\
& & \qquad + \frac{n_{00}}{N} \big|\epsilon_1 (\theta_1^*) 
- \tilde{\epsilon}_1(\tilde{\theta}_1^*) \big|
+ \frac{n_{10}}{N} \gamma
\big),
\end{eqnarray*}
where we further have that 
\begin{eqnarray*}
|\fp_1(\theta_1^*)-\fp_1(\tilde{\theta}_1^*)|
&=& |F_{01}(\theta_1^*) - F_{01}(\tilde{\theta}_1^*)|
\\
&\leq& \hat{f}_{01} |\theta_1^* - \tilde{\theta}_1^*|
\\
&\leq& \hat{f}_{01} M_{11} \gamma/2,   
\end{eqnarray*}
and 
\begin{eqnarray*}
\big|\epsilon_1 (\theta_1^*) 
- \tilde{\epsilon}_1(\tilde{\theta}_1^*) \big|
&\leq& \big|\epsilon_1 (\theta_1^*) 
- \tilde{\epsilon}_1(\theta_1^*) \big| \\
& & \qquad +\big|\tilde{\epsilon}_1(\theta_1^*) 
- \tilde{\epsilon}_1(\tilde{\theta}_1^*) \big|
\\
&\leq& \big|F_{00}(F^{-1}_{10}(F_{11}(\theta_1^*))) \\
& & \qquad - F_{00}(F^{-1}_{10}(F_{11}(\theta_1^*)+\gamma)) \big| \\
& & \qquad + \hat{\epsilon}_1^\prime M_{11} \gamma/2
\\
&=& (\hat{f}_{00} M_{10} + \hat{\epsilon}_1^\prime M_{11}/2) \gamma . 
\end{eqnarray*}
Here, $\hat{\epsilon}_1^\prime = \max \tilde{\epsilon}_1^\prime$ is the maximum of the derivative of $\tilde{\epsilon}_1$.  
Combining all the terms in front of $\gamma$ gives us the desired upper bound. 
\end{proof}

Theorem 3 quantifies the decrease in accuracy loss (\ie the improvement in accuracy) when we allow a gap of true positive rates between two groups (\ie relaxation from the perfect EOp condition).

Repeating the analysis for the perturbed PE condition, we can obtain a similar bound for the changes in the accuracy loss function. We omit the details here in the interest of space. 

\section{Convergence Analysis of GSTAR}
\subsection{GSTAR as Nonlinear Least Squares Problem}

Our objective function and the optimization solution algorithm belong to the family of \textbf{Gauss-Newton algorithm} to solve Nonlinear Least Squares Problem (NLSP). To specify, NLSP is to solve 
\[
\min_{\boldsymbol{\theta}} ||r(\boldsymbol{\theta})||^2_2, 
\]
where the decision variables, $\boldsymbol{\theta}$, is an $n$-dimensional real vector and the objective function $r$ is an $m$-dimensional real vector function of $\boldsymbol{\theta}$. Connecting to our setting and using two groups as an example, our decision variables is the two-dimensional vector $\boldsymbol{\theta}=(\theta_0, \theta_1)$ for group 0 and group 1, and our objective function is the following 2-dimensional real vector function:
\[
r(\boldsymbol{\theta}) = \big( r_1(\boldsymbol{\theta}), r_2(\boldsymbol{\theta}) \big)
\]
with 
\begin{eqnarray*}
r_1(\boldsymbol{\theta}) = r_1(\theta_0, \theta_1) &=& \frac{n_{01}}{N}\fp_1(\theta_1) + \frac{n_{11}}{N}\fn_1(\theta_1) \\
& & \qquad + \frac{n_{00}}{N}\fp_0(\theta_0) + \frac{n_{10}}{N}\fn_0(\theta_0), 
\\
r_2(\boldsymbol{\theta}) = r_2(\theta_0, \theta_1) &=&
\sqrt{\lambda}\left(\tp_1(\theta_1) - \tp_0(\theta_0)\right) 
\end{eqnarray*}
when taking the EOp constraint. 
The $L_2$ norm $||r(\boldsymbol{\theta})||^2_2 = r_1(\boldsymbol{\theta})^2 + r_2(\boldsymbol{\theta})^2$ recovers the objective function in Equation (2) in the main paper.

A classic family of algorithms to solve NLSP is the Gauss-Newton Method. The main idea is to convert the nonlinear optimization problem to a linear least square problem using Taylor expansion. That is, the parameter values are calculated in an iterative fashion with 
\[
\theta_{j} \approx \theta_{j}^{k+1} = \theta_{j}^{k} +\Delta_j,
\] 
in the $k$-th iteration number, with the vector of increments $\Delta=\{\Delta_j\} = \{ \theta_{j}^{k+1} - \theta_{j}^{k} \}$ (also known as the shift vector). We linearize each component in the $f$ function to a first-order Taylor polynomial expansion as 
\begin{equation}
r_i (\boldsymbol{\theta}) \approx r_i(\boldsymbol{\theta}^k) + \sum_{j} \frac{\partial r_i(\boldsymbol{\theta}^k)}{\partial\theta_j} \Delta_j 
\label{eq-sup:Taylor}
\end{equation}
with $\boldsymbol{\theta}^k =(\theta_0^{k}, \theta_1^k)$. Plugging this linearized equation into the objective function, we get the usual least square problem. Then, the optimal solution can be obtained as 
\begin{equation}
\Delta = - (J^T J)^{-1} J^T f(\boldsymbol{\theta}^k), 
\label{eq-sup:NLSP-solution}
\end{equation}
where $J=\{J_{ij}\}$ with $J_{ij} = \{\frac{\partial r_i(\boldsymbol{\theta})}{\partial\theta_j}\}$ is the Jacobian. Note that in the GSTAR algorithm, we ignore the terms for $j\neq i$ in the Taylor expansion~(\ref{eq-sup:Taylor}). Thus, in calculating $J^T J$, we only kept the diagonal terms 
\[
\left(\frac{\partial  r_1(\boldsymbol{\theta})}{\partial\theta_j}\right)^2
+ \left(\frac{\partial  r_2(\boldsymbol{\theta})}{\partial\theta_j}\right)^2
\]
for $j=0,1$. Plugging in the form of $r_1$ and $r_2$ as specified above, we achieve the solution provided in \eqref{eq-sup:update_delta}.

\subsection{Convergence Property for Gauss-Newton Algorithm}
There is a long history of studying the convergence property of the Gauss-Newton algorithm, \eg see \cite{gratton2007approximate}. The convergence of the algorithm is generally not guaranteed, \eg if the initial solution is far from the true optimal or $J^T J$ is ill-conditioned. In other words, the convergence of the algorithm heavily depends on the density estimation $f(\cdot)$. We state the following sufficient conditions from~\cite{gratton2007approximate} to guarantee the convergence of the algorithm. 
The following assumptions are
made in order to establish the theory. 
\begin{itemize}
    \item A1. There exists $\theta^*$ such that $J^T(\theta^*)r(\theta^*) = 0$;
    \item A2. The Jacobian at $\theta^*$ has full rank.
\end{itemize}
We state Theorem~4 from \cite{gratton2007approximate} on the sufficient conditions for convergence.
\setcounter{theorem}{3}
\begin{theorem}
Assume that the estimated density function $f(\cdot)$ satisfy assumptions A1 and A2. Further, $f(\cdot)$ satisfies that 
\[
||Q(\theta^k)(J^T J)^{-1}(\theta^k)||_2 \leq \eta 
\]
for some constant $\eta\in[0,1)$ for each iteration $k$, where $Q(\theta)$ denotes the second order terms $\sum_i r_i(\theta) \nabla^2 r_i(\theta)$. Then as long as the initial solution is sufficiently close to the true optimal with $||\theta^0 -\theta^*||_2 \leq \epsilon$, the sequence of Gauss-Newton iterates $\{\theta^k\}$ converges to $\theta^*$. 
\end{theorem}

\subsection{Protection against Divergence}

It is known that for general function $f(\cdot)$ such as estimates from a neural network, the above sufficient conditions that guarantee convergence do not necessarily hold. As a result, protection against divergence is essential. In our numerical experiments, we adopt a commonly used, simple protection, the shift-cutting method. That is, we to reduce the length of the shift vector $\Delta$ by a fraction $\eta$. In other words, the update becomes
\[
\theta_j^{k+1} = \theta_j^k + \eta \Delta_j.  
\]

\section{Near Optimality of GSTAR}
Here, we show regarding on how the accuracy of $h$ affects the accuracy of $\hat{Y}$. 
Following the proof of Theorem 5.6 of Hardt \etal ~\shortcite{hardt2016equality}, we provide the following near optimality results for our method. 


Before we prove the theorem, we first state the results from Lemma 5.5 proved in Hardt et al. (2016), which will be used in our proof. 

\begin{lemma} [Restatement of Lemma 5.5 in Hardt et al. \shortcite{hardt2016equality}]
Let $R, R^\prime \in [0,1]$ be two random variables in the same probability space as $A$ and $Y$. Then, for any point $p$ on a conditional ROC curve of $R$, there is a point $q$ on the corresponding ROC curve of $R^\prime$ achieving the same threshold such that
$$|| p - q||_2 \leq \sqrt{2} d_K(R, R^\prime), $$
\begin{equation}
\begin{split}
d_K (R, R^\prime) &= \max_{a,y} \sup_t | Pr(R > t |A=a, Y=y) \\
   & \qquad - Pr(R^\prime > t |A=a, Y=y) |.     
\end{split}
   \label{eq-sup:kol}
\end{equation}
\end{lemma}

\begin{proof}
Similar to Hardt et al. \shortcite{hardt2016equality}, we focus on proving this theorem for equalized odds. The case of equal opportunity is analogous.
The optimal classifier $Y^*$ corresponds to a point $p^*$. 
Under the equalized odds condition, our algorithm essentially finds the intersection point, $q$, of the two conditional ROC curves of $\hat{R}_h$ for $a=0$ and $a=1$. Then directly applying the above lemma, we get that  
$$ || p^* - q ||_2 \leq \sqrt{2} d_K(R, R^\prime) .$$
The rest follows the same argument as in Theorem 5.6 of Hardt et al. \shortcite{hardt2016equality}. That is, by assumption on the loss function, there is a vector $v$ with $||v||_2 \leq \sqrt{2}$ such that $\mathbb{E}[\ell(\hat{Y}_h, Y)] = \langle v,q \rangle$ and $\mathbb{E}[\ell(Y^*, Y)] = \langle v, p^* \rangle$. Applying Cauchy-Schwarz, we get 
\begin{eqnarray*}
         \mathbb{E}[\ell(\hat{Y}_h, Y)] - \mathbb{E}[\ell(Y^*, Y)] &=& \langle v, q-p^* \rangle \\
&\leq& ||v||_2 \cdot || p^* - q ||_2 \\
&\leq&  2 d_K(R, R^\prime).
\end{eqnarray*}
\end{proof}

\begin{remark}
In Hardt et al. \shortcite{hardt2016equality}, the point $q$ from their algorithm under equalized odds condition is the intersection point between two \textit{line segments}, not the two ROC curves as in our paper. That is, assume without loss of generality that the first coordinate of $q_1$ (for group $a=1$) is greater than the first coordinate of $q_0$ (for group $a=0$) on the ROC curve plane; and that all points $p^*, q_0, q_1$ lie above the main diagonal. Then $q \in L_0 \cap L_1$ from their algorithm, where $L_0$ is the line segment between $q_0$ and the point $(1,1)$, and $L_1$ is the line segment between the point $(0,0)$ and $q_1$. As a result, in proving their Theorem 5.6, they need to show that $q$ from this construction satisfies $|| p^* - q ||_2 \leq 2 d_K(R, R^\prime).$ However, because the point $q$ from our algorithm lies on the ROC curve, we can directly apply the results from the lemma. This difference is further illustrated in figure \ref{fig:path} below, where the purple pentagram corresponds to $q$ found by our algorithm, and the green cross corresponds to $q$ from their algorithm. The figure shows the intersection points found from our algorithm versus Hardt et al.

Moreover, the requirement for achieving the near optimality in our method (our Theorem 5) and in Hardt et al. (their Theorem 5.6) is the same. That is, the closeness between the conditional densities is required, not just the conditional probability estimates.

To specify, the closeness requirement in Hardt et al. based on conditional Kolmogorov distance is shown in Equation \eqref{eq-sup:kol}, where $R\in[0, 1]$ and $R' \in [0, 1]$ are real-valued scores, \ie  two regressors. Note that the distance is taking $\sup$ over all $t \in [0,1]$, so this condition requires the entire conditional density curves from $R$ and $R'$  to be close, not just close at some given threshold $t$.

Next, the near optimality of Hardt et al. (their Theorem 5.6) shows:
$$ \mathbb{E}[\ell(\hat{Y}_h, Y)] \leq 
 \mathbb{E}[\ell(Y^*, Y)] + 2 \sqrt{2} d_K(\hat R, R^*),$$ 
where $R^* \in [0, 1]$ is the Bayes optimal regressor and $\hat R \in [0, 1]$ is a regressor whose density is estimated. In fact, the distribution function of their 
 corresponds to the score function $\sigma(h)$ in our paper, where $h$ is the logit and  $\sigma(\cdot)$ is the softmax function.

Seeing this connection, we stress that the closeness requirement in our result is the same as in Hardt et al., and that the near optimality of our algorithm follows:
$$ \mathbb{E}[\ell(\hat{Y}_h, Y)] \leq 
 \mathbb{E}[\ell(Y^*, Y)] + 2 d_K(\hat R, R^*),$$ 
 where $R^* \in [0, 1]$ is the Bayes optimal regressor as given in Hardt et al., and $\hat R \in [0, 1]$  comes from our estimated density, \ie, the distribution of $\hat R_h$  comes from by applying softmax function $\sigma(\cdot)$ on logit $h$.

\end{remark}

\begin{figure}
    \centering
    \includegraphics[width = 1\linewidth]{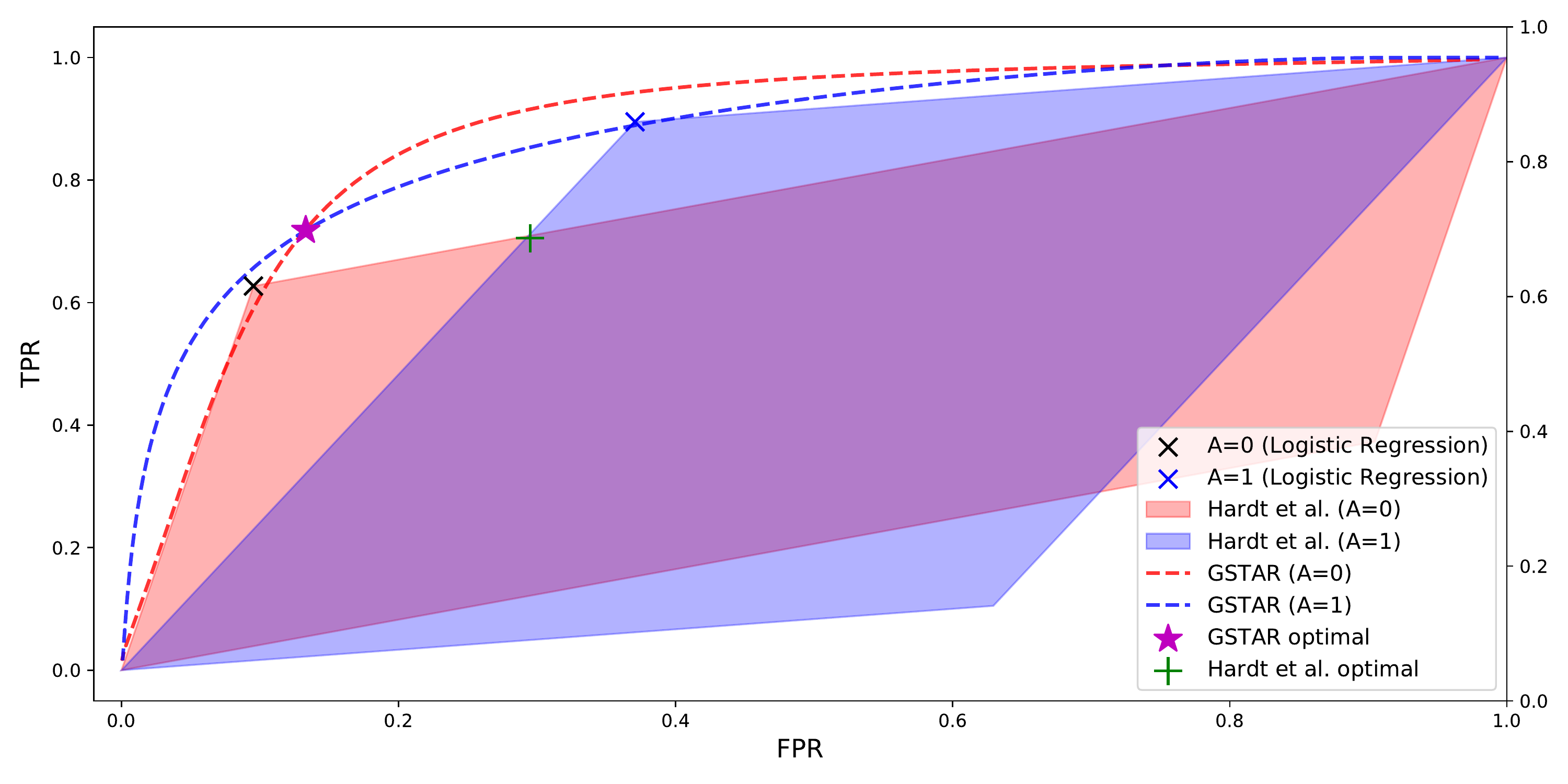}
    \caption{Comparison of optimal point of Hardt \etal and GSTAR. Given the ROC curve of each protected group, ours (magenta star) achieves better optimum than that of Hardt \etal (green cross), as ours has higher TPR and lower FPR.}
    \label{fig:path}
\end{figure}
\section{Experimental Details}
\subsection{Comparing Methods}
We compared our method with multiple state-of-the-art methods to verify our work. The details about the comparing methods are as below:
\begin{itemize}
    \item \textbf{Learning fair representations for kernel models} (abbreviated as FGP)~\cite{tan2020learning}: a pre-processing method to learn representation focusing on kernel-based models. The fair model that satisfies certain fairness criterion is obtained by Bayesian learning from fair Gaussian process (FGP) prior.
    \item \textbf{Fairness confusion tensor} (abbreviated as FACT)~\cite{kim2020model}: a post-processing model that minimize the least-squares accuracy-fairness optimality problem based on confusion tensor.
    \item \textbf{Adversarial de-biasing} (abbreviated as AdvDeb)~\cite{zhang2018mitigating}: an in-processing model that mitigates the conflicting gradient directions in utility and fairness objectives by projecting one gradient to another to remove the opposite direction.
    \item \textbf{Calibrated equal odds post-processing} (abbreviated as CEOPost)~\cite{pleiss2017fairness}: a post-processing method that minimizes the disparity in the predicted probability to the preferred class among different sensitive groups, while maintaining the calibration condition in a relaxed condition.
    \item \textbf{Equality of opportunity in supervised learning} (abbreviated as Eq.Odds)~\cite{hardt2016equality}: a post-processing method that learns the threshold to yield the equalized odds/opportunity between different demographic by exploring the intersection of achievable true positive rates and false positive rates. 
    \item \textbf{Learning adversarially fair and transferable representations} (abbreviated as LAFTR)~\cite{madras2018learning}: a fair representation learning model that adopts fairness metrics as the adversarial objectives and analyze the balance between utility and fairness.
    \item \textbf{Baseline}: For CelebA dataset, we use ResNet50~\cite{he2016deep} as a reference because we input second last layer (2048 features) of ResNet to all methods. For other tabular datasets, logistic regression is used as all other methods except for FGP and LAFTR are based on logistic regression.
\end{itemize}

\subsection{Evaluation Metrics}
In the experiments, we evaluate the methods on four fairness and two performance measures.
Four fairness metrics are as below:
\begin{itemize}
    \item \textbf{Equal Opportunity} (abbreviated as EOp)~\cite{hardt2016equality} : This measures absolute difference of favorable prediction given positive label.
    $$|P(\hat Y=1|Y=1, A=1)-P(\hat Y=1|Y=1, A=0)|. $$

    \item \textbf{Equalized Odds} (abbreviated as EOd)~\cite{hardt2016equality} : This measures the difference between the probability given the true labels.
    \begin{eqnarray*}
    \begin{aligned}
        |P(\hat Y=1|Y=1, A=1)-P(\hat Y=1|Y=1, A=0)| + \\
        |P(\hat Y=1|Y=0, A=1)-P(\hat Y=1|Y=0, A=0)|. 
    \end{aligned}
    \end{eqnarray*}
    
    \item \textbf{Balanced Accuracy Difference} (abbreviated as BD) : This measures difference between balanced accuracy between the groups.
    \begin{eqnarray*}
            \begin{aligned}
            &\Big|P(\hat Y=1|Y=1, A=1)+P(\hat Y=0|Y=0, A=1)\Big| \\
            & - \Big|P(\hat Y=1|Y=1, A=0)+P(\hat Y=0|Y=0, A=0)\Big|. 
            \end{aligned}
    \end{eqnarray*}
    If BD and EOd has the same value, it indicates that both TPR and TNR are higher in a certain sensitive group. However, if the gap between the two terms is large, we can interpret as the classifier is more biased as a group with higher TPR has lower TNR. This is more unfair as a sample from the privileged group is more likely to be falsely and correctly predicted as positive output. EOp is a partial measure of EOd as it only measures the difference from a favorable class. 
    
    \item \textbf{Absolute (1 - Disparate Impact)} (abbreviated as 1-DIMP)~\cite{barocas2016big} : This measures
    ratio of the probability of the favorable prediction given a protected group.
    \[
    \left|1 - \frac{P(\hat Y = 1|A=1)}{P(\hat Y = 1|A=0)}\right|.
    \]
\end{itemize}

We evaluate performance of the methods with two metrics.
\begin{itemize}
    \item \textbf{Balanced Accuracy} (abbreviated as BA) : This measures average between true positive rate and true negative rate. Compared to the traditional accuracy, this measure effectively shows the whether the classifier is focusing on the performance of a certain class when the dataset is unbalanced.
    \[
    \frac{1}{2}\left(P(\hat Y=1|Y=1)+P(\hat Y=0|Y=0)\right).
    \]

    \item \textbf{Accuracy} (abbreviated as ACC) : This measures traditional classification accuracy of the method.
\end{itemize}

\subsection{Experimental Setup}
For experimental setup, all comparing methods apply EOd as the fair constraint if fairness constraint is selectable, thus we compare them via EOd in Figure 2 in the main paper. Both the Pareto frontier from GSTAR and FACT are derived based on EOd constraint for a fair comparison. We follow the setup in Section G.3 of the FACT \cite{kim2020model} to report their method, which does not require $\lambda$. 

For GSTAR, we estimate $f_{ya}$  and optimize $\theta_a$ from the training data, and report evaluation results (with the $\theta_a$ learned from training data) on the testing data. We use the same $\lambda$ for multiple fairness constraints for simplicity, but $\lambda$ can be introduced individually.
Our method is optimized with $\lambda$ in the range of $[10^{-1}, 10^{4}]$ with alternating optimization method.

To find estimated distribution $f_{ya}$, we consider gamma, Student's t, and normal distribution as the candidates for the experiments reported in the main paper, and select the one that has the maximum likelihood with the output distribution. Without loss of generality, this can be generalized non-parametric density estimation such as kernel density estimation to fit more complicated distribution. More experiments with complicated distribution estimation is in Section \ref{sec:kde} in the supplementary.

Figure 2 illustrates Pareto frontiers with 5 points of different $\lambda$ values in $[10^{-2}, 10^7]$ with equal logspace. 
To visualize the plots, we sweep hyperparameters (e.g, weights for each term in the objective function) for comparing methods.
Figure 3 takes $\lambda$ or hyperparameter values from the upper-right point of the Pareto frontiers in Figure 2, which indicates the best trade-off for each method. Figure 3 paper presents the 5 runs with the setup chosen based on the Pareto frontier to show the consistency of the performance of each model.

All experiments are implemented with Pytorch framework on i9-9960X CPU and a Quadro RTX 6000 GPU.

\subsection{Dataset Description}
We evaluate the methods on four fairness datasets. The goal for all datasets is binary classification on binary sensitive feature. The details of the datasets are as below:

\begin{itemize}

    \item \textbf{CelebA image dataset}\footnote{\url{http://mmlab.ie.cuhk.edu.hk/projects/CelebA.html}}~\cite{celeba}: The data consists of 202,599 face images in diverse demographics. The images are annotated with 40 attributes (face shape, skin tone, smiling, etc.). Similar to Quadrianto \etal~\cite{attractive2019}, the goal is to predict whether a person in the image is attractive or not. The feature \emph{sex} is used as the sensitive feature. 

    \item \textbf{Adult} dataset from the UCI repository~\cite{adult1996} contains 48,842 instances described by 14 features (workclass, age, education, sex, race, \etc) with the goal of the income prediction whether a person's income exceeds 50K USD per year. The feature \emph{sex} is used as the sensitive feature.

    \item \textbf{COMPAS}\footnote{\url{https://github.com/propublica/compas-analysis}}(Correctional Offender Management Profiling for Alternative Sanctions) dataset includes 6,167 samples described by 401 features with the target of recidivism prediction with the label showing if each person gets rearrested within two years. The feature \emph{race} is used as the sensitive feature for this dataset.

    \item \textbf{German} credit dataset from the UCI repository~\cite{Dua2019} contains 1,000 samples described by 20 features. The goal is to predict the credit risks. The feature \emph{sex} is used as the sensitive feature.

\end{itemize}
 All data is split as 70\% for training and 30\% for testing.
 
 \begin{table}[!t]
    \centering
    \begin{tabular}{c||c|c|c|c}
        \hline
        \multicolumn{5}{c}{CelebA} \\ \hline\hline
        Model & GSTAR & FGP & FACT & CEOPost \\\hline
        Time & 0.287 &  - & 0.067 & 0.077 \\ \hline
        Model & DIR & Eq.Odds & LAFTR & AdvDeb\\\hline
        Time  & 183.20 & 0.062 & 107.04(min) & 303.15\\\hline
        \end{tabular}
    \hfill
    \begin{tabular}{c||c|c|c|c}
            \hline        
        \multicolumn{5}{c}{Adult} \\ \hline\hline
        Model & GSTAR & FGP & FACT & CEOPost \\\hline
        Time & 0.29 & 51.28 & 0.055  & 25.61 \\ \hline
        Model & DIR & Eq.Odds & LAFTR & AdvDeb\\\hline
        Time  & 168 & 0.037 & 53.04(min) & 102.00\\\hline
    \end{tabular}
        
    \begin{tabular}{c||c|c|c|c}
    \hline
    \multicolumn{5}{c}{Compas} \\ \hline\hline
    Model & GSTAR & FGP & FACT & CEOPost \\\hline
    Time & 0.292 & 43.74 & 0.035 & 8.3 \\ \hline
    Model & DIR & Eq.Odds & LAFTR & AdvDeb\\\hline
    Time  & 123.20 & 0.034 & 57.04(min) & 15.45\\\hline
    \end{tabular}
        
    \begin{tabular}{c||c|c|c|c}
    \multicolumn{5}{c}{German} \\ \hline\hline
    Model & GSTAR & FGP & FACT & CEOPost \\\hline
    Time & 0.271 & 7.08 & 0.0257 & 2.64 \\ \hline
    Model & DIR & Eq.Odds & LAFTR & AdvDeb\\\hline
    Time  & 1.68 & 0.034 & 56.51(min) & 2.17\\\hline
    \end{tabular}
\caption{Computational time (in seconds) for all comparing fairness methods for each dataset.}
\label{tab:time}
\end{table}

\subsection{Computational Cost}
In \autoref{tab:time}, we describe the computational time for each method on each dataset.
By introducing estimated PDF functions for post-processing, we outperform other methods except Eq.Odds~\cite{hardt2016equality} and FACT~\cite{kim2020model}.
As they both only utilize the entries of the confusion matrix to find optimal mixing rate in their methods, they have less computation than ours.
However, as we discussed in the main paper, we explore better feasible region than theirs by group-specific thresholding that results better in both fairness and performance by sacrificing little efficiency, yet outperforms most of the other works.

\begin{figure*}[!t]
    \centering
    \includegraphics[width =\linewidth]{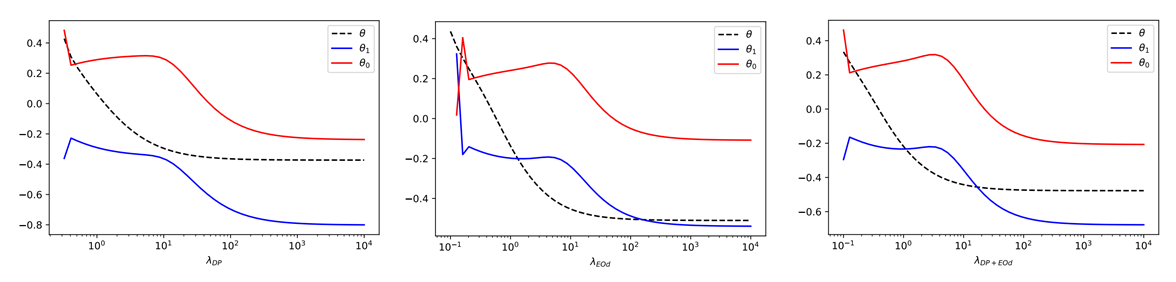}
    \caption{Trend of converged $\theta$ values based on the variation of weight $\lambda$. Dashed line indicates single threshold version and $\theta_a$ indicates threshold for $a$ group.}
    \label{fig:single_thres}
\end{figure*}
\begin{figure}[!t]
    \centering
    \includegraphics[width = 1.1\linewidth, height=.55\linewidth]{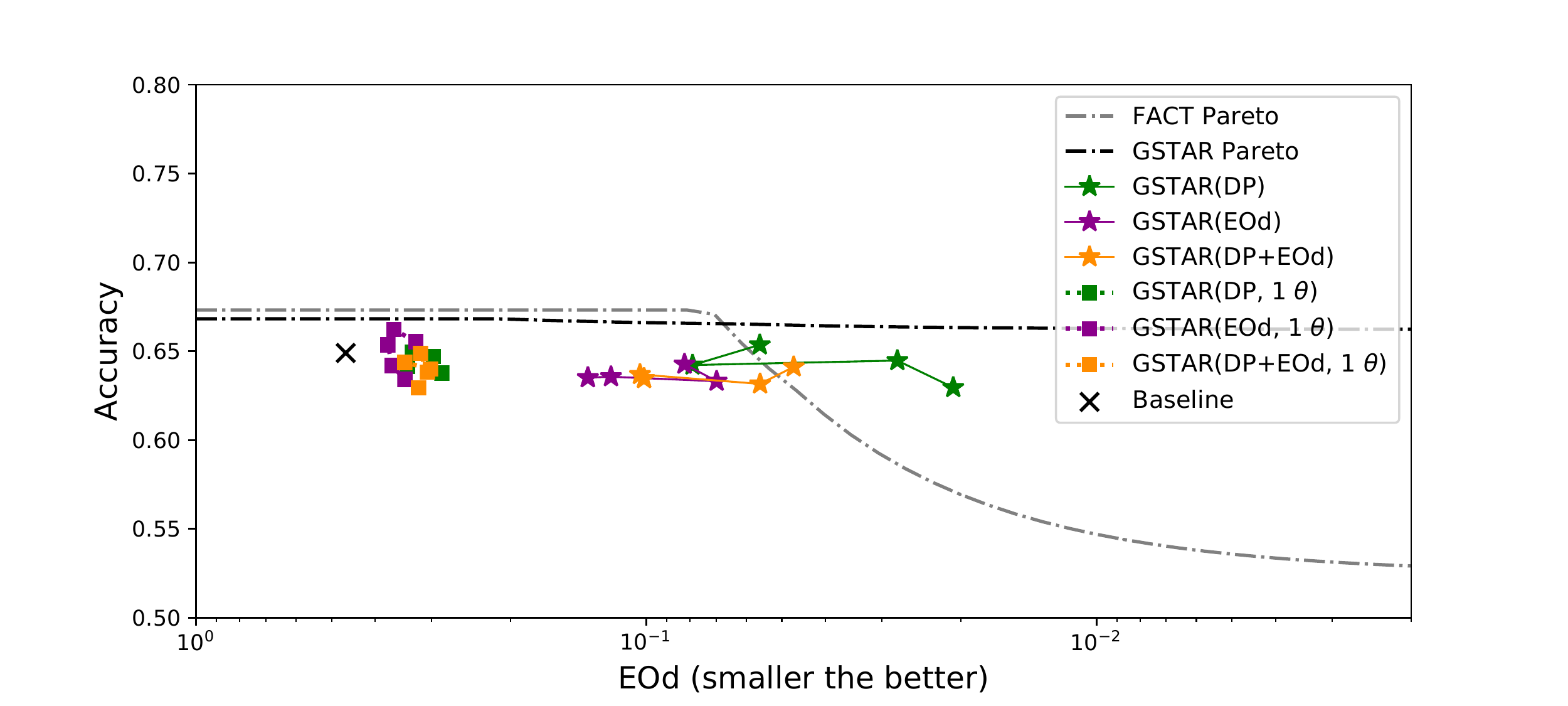}
    \caption{Comparison of single threshold (squares, $1\theta$) and group-aware threshold method (stars) on Pareto frontier. The result suggests group-aware threshold greatly improve fairness with comparable accuracy.}
    \label{fig:single_thres_front}
\end{figure}

\begin{figure}[!t]
    \centering
    \begin{subfigure}[b]{0.45\textwidth}
     \includegraphics[width = .95\linewidth, height=.6\linewidth]
     {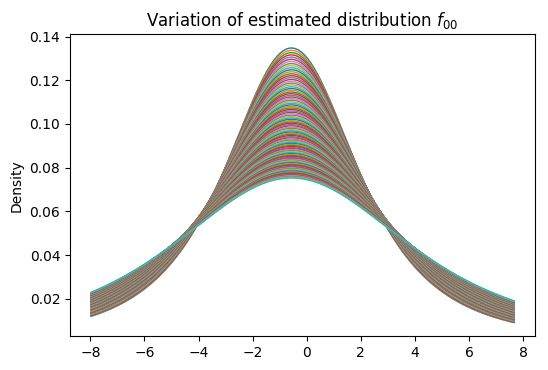}
    \caption{Variation of estimated distribution of $f$ by the noise factor $\alpha$.}
    \label{fig:variation}   
    \end{subfigure}
    \hfill
    \begin{subfigure}[b]{0.49\textwidth}
    \includegraphics[width = 1\linewidth, height=.55\linewidth]
    {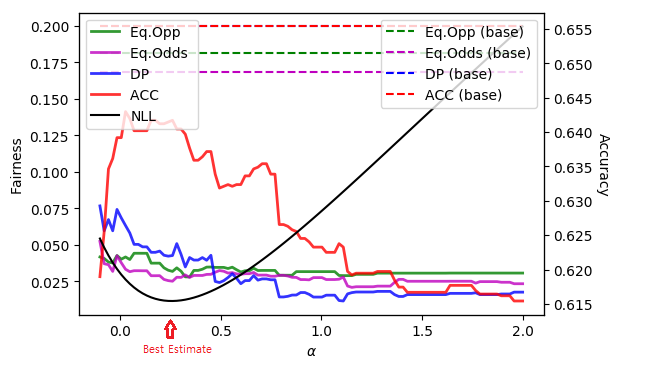}
    \caption{The influence of the noise factor $\alpha$ and NLL of corresponding estimated distribution to the performance and fairness of GSTAR.}
    \label{fig:noise}    
    \end{subfigure}
    \caption{Variation of estimated distributions by the noise $\alpha$ and its impact on the performance of GSTAR.}
    \label{fig:dist_est}   
\end{figure}





\section{Auxiliary Experiments}
\subsection{GSTAR with single threshold}
We conduct experiments on COMPAS dataset to evaluate GSTAR with a single adaptive thresohld. Figure \ref{fig:single_thres_front} presents the trend of fairness-accuracy trade-off of two versions of GSTAR based on $\lambda$ values. Comparing with the baseline ($\theta = 0$), we observe that even with a single threshold in GSTAR (1$\theta$ in the legend), the adaptive threshold helps to improve the fairness with comparable accuracy. However the improvement is not as significant as that of the group-wise version because it is impossible to achieve perfect fairness with a single threshold as the intersection of  $f_{1a}$ and $f_{0a}$ differs by $a$. Figure \ref{fig:single_thres} shows the trend of learned $\theta$  based on $\lambda$ values. We see a single threshold version (black) lies between two thresholds of group-aware GSTAR in most cases. This implies that the single threshold converges to some point that gives up some of the fairness.

\begin{table*}[!t]
    \centering
    \begin{tabular}{c|c|c|c|c|c|c}
         & ACC (train) & DP (train) & EOd (train) & ACC (test) & DP (test) & EOd (test) \\\hline
         GSTAR (DP) & 0.679 & 0.001 & 0.089 & 0.639 & 0.017 & 0.018 \\
         GSTAR (EOd) & 0.714 & 0.071 & 0.030 & 0.0643 & 0.032 & 0.034 \\
         GSTAR (DP + EOd) & 0.705 & 0.050 & 0.030 & 0.641 & 0.027 & 0.025 \\\hline
    \end{tabular}
    \caption{Comparison of fairness and performance measure in training and testing set with different fairness constraints.}
    \label{tab:train-test}
\end{table*}
\begin{figure}[!t]
	\centering
	\begin{minipage}[t]{0.49\textwidth}
		\centering
		\includegraphics[width=1.0\linewidth, height =0.48\linewidth]{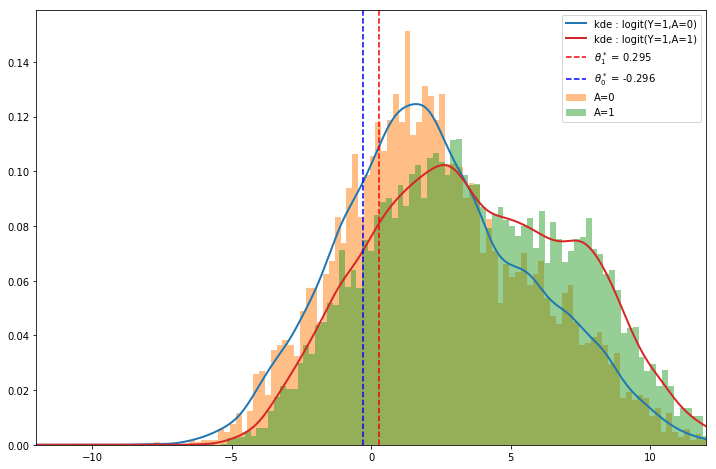}
	\end{minipage}
    \hfill
	\begin{minipage}[t]{0.49\textwidth}
		\centering
		\includegraphics[width=1.0\linewidth, height =0.48\linewidth]{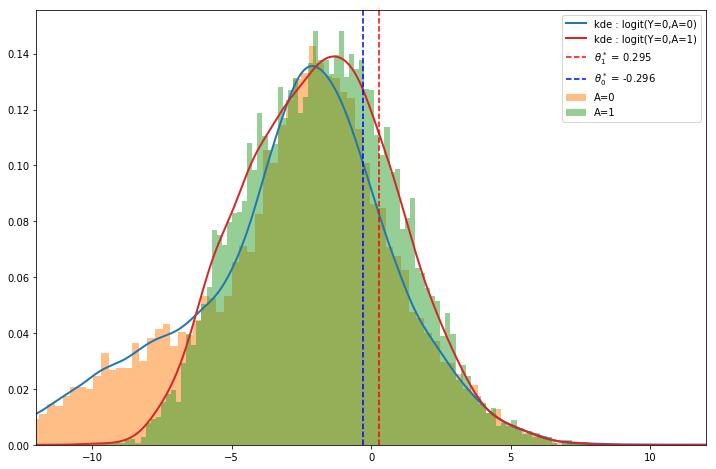}
	\end{minipage}
	\caption{Distribution of synthetic dataset and its kernel density estimation.}
	\label{fig:kde}
\end{figure}

\subsection{Quality of Estimated Distribution}
The performance of GSTAR relies on the quality of estimated distribution. 
For the benchmark datasets, we empirically found that the distribution of logits resembles some parametric distributions. 
Thus, we estimate the distribution with generally used parametric distributions such as Student's t-distribution by measuring the negative log-likelihood (NLL) in the training data. Note that GSTAR can be extended to a wide range of other distributions, even non-parametric distributions.

For further analysis, we add new experiments by sweeping the parameters of parametric distribution to see the effect of the estimation quality. In COMPAS dataset, the best estimate (\ie smallest NLL) of group ($y=0, a=0$) with Student's t-distribution has parameters of df = 2.235, loc = -0.567, scale = 0.756 based on scipy package. To generate variations as in Figure \ref{fig:variation} of distributions with varying estimation qualities, we add noise $\alpha \in [-0.1, 100]$ to the scale of this distribution.

In figure \ref{fig:noise}, we illustrate the trend of NLL (black), fairness violation (the lower the better), and accuracy (the higher the better) with varying noise ($\alpha$, x-axis). The color of lines follows the main paper. Dashed lines indicate the quantity of baseline model ($\theta =0$). From this, we observed that the accuracy is the most sensitive to the change of estimation quality, while fairness is relatively stable.

However, for our experiments, we assume the estimation is reliable and the guarantee on the estimation reliability is beyond our focus of this paper.

\begin{figure*}[!t]
	\centering
	\begin{minipage}[t]{0.33\textwidth}
		\centering
		\includegraphics[width=1.05\linewidth]{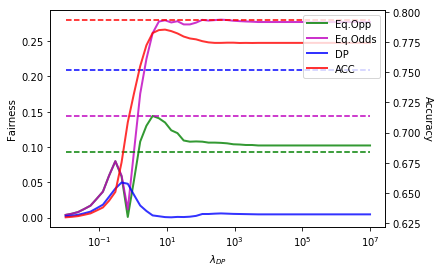}
		\subcaption{$\LL_{fair} = \LL_{fair}^{DP}$}
	\end{minipage}
    \hfill
	\begin{minipage}[t]{0.33\textwidth}
		\centering
		\includegraphics[width=1.05\linewidth]{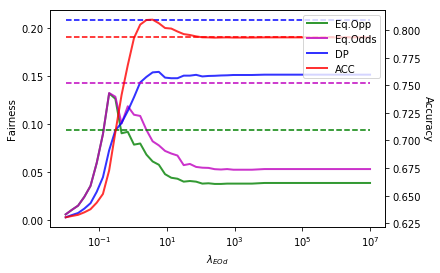}
		\subcaption{$\LL_{fair} = \LL_{fair}^{EOd}$}
	\end{minipage}
    \hfill
	\begin{minipage}[t]{0.33\textwidth}
		\centering
		\includegraphics[width=1.05\linewidth]{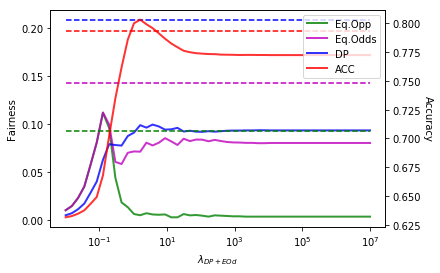}
		\subcaption{$\LL_{fair} = \LL_{fair}^{DP+EOd}$}
	\end{minipage}
	\caption{Trend of performance and fairness measure by the change of $\lambda$ values.
	Color of the lines indicates the measure of performance and fairness as in the legend. Solid lines indicate GSTAR results and dotted lines indicate  baseline ($\theta = (0,0)$) respectively.
	It is lower the better for fairness and higher the better for accuracy.}
	\label{fig:kde_perf}
\end{figure*}
\begin{figure*}[!t]
	\centering
	\begin{minipage}[t]{0.33\textwidth}
		\centering
		\includegraphics[width=.9\linewidth, height = 0.65\linewidth]{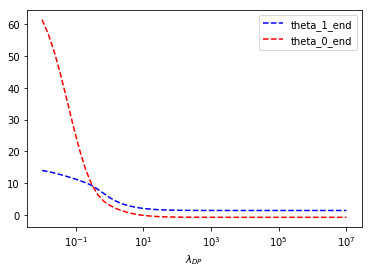}
		\subcaption{$\LL_{fair} = \LL_{fair}^{DP}$}
	\end{minipage}
    \hfill
	\begin{minipage}[t]{0.33\textwidth}
		\centering
		\includegraphics[width=.9\linewidth, height = 0.65\linewidth]{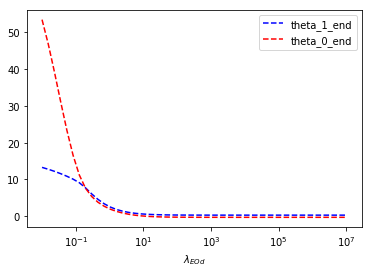}
		\subcaption{$\LL_{fair} = \LL_{fair}^{EOd}$}
	\end{minipage}
    \hfill
	\begin{minipage}[t]{0.33\textwidth}
		\centering
		\includegraphics[width=.9\linewidth, height = 0.65\linewidth]{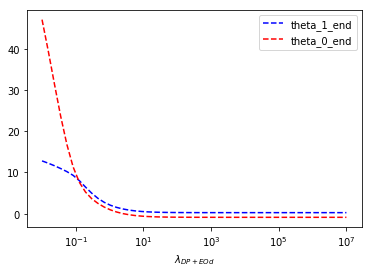}
		\subcaption{$\LL_{fair} = \LL_{fair}^{DP+EOd}$}
	\end{minipage}
	\caption{Trend of converged group-aware threshold $\theta$ achieved by GSTAR.}
	\label{fig:kde_theta}
\end{figure*}

\subsection{Interpretation of Results on COMPAS}
In COMPAS in Figure 3(c), we observe improvements in total fairness violation with multiple fairness constraints employed. We deduce this could happen due to: 1) generalization of the estimated distribution from training data to testing data; 2) difference in the training and testing data distributions. For the training set, we achieve better fairness violation on the model with a single constraint, compared to the multi-constrained version or other single-constrained versions. In the training set of COMPAS data, we have the results as in the Table \ref{tab:train-test}.

\subsection{Complicated Distribution Estimation with Kernel Density Estimation}
\label{sec:kde}
We can generalize our density estimation to non-parametric by kernel density estimation (KDE) method.
Given the logit distribution $h(X)$, we build a histogram with $B$ bins. Denote $T_b$ as the mean logit value of $b$-th bin and $w_b$ as normalized weight indicates how many samples belong to $b$-th bin, where $b \in \{1, \cdots, B\}$ and $\sum_b w_b = 1$.
Then our kernel density estimator of distribution $h(X)$ is 
\begin{equation}
    f(x) = \sum_b w_b  K(x - T_b),
    \label{eq-sup:kde}
\end{equation}
where $K$ is kernel function and we employ normal distribution with standard deviation as 0.5.
As non-parametric density estimation of $h(x)$ can be expressed as linear combination of parametric distributions, we can easily apply the optimization step demonstrated in the \autoref{sec:optim_method}.

To validate the KDE method for GSTAR, we generate synthetic data that each logit distribution $h_{ya}$ consist of mixture of three gaussian distributions with additional standard normal noise. Specifically, each distribution is configured with mean $\mu$, variation $\sigma^2$, and weight $w$ and number of samples $n$ as in \autoref{tab:syn}.
For example, we generate the samples from a group ($Y=0, A=0$) by sampling $n_{00}$ samples $x$ on $h_{00}$ and add noise $\N(0, 1^2)$ as below:
$$h_{00} =  \sum_{i \in \{0,1,2\}} w_{00}^{(i)} \cdot \N(\mu_{00}^{(i)}, \sigma_{00}^{(i)^2}), $$
$$l^{(k)} \sim h_{00}, \quad \epsilon^{(k)} \sim \N(0, 1^2),$$
$$x^{(k)} := l^{(k)} + \epsilon^{(k)}, \quad k \in \{1, \cdots, n_{00}\}$$
where $i$ is the index of gaussian distributions in \autoref{tab:syn} and $k$ is the index of sampling instance.

\begin{table}[!h]
    \centering
    \small
    \begin{tabular}{c|c|c|c|c}
                 & $\mu$        &   $\sigma^2$    &    $w$   &   $n$   \\\hline
       $h_{00}$  & [-7.0, -2.0, 1.1]  &  [3.0, 1.5, 2.0]  & [0.3, 0.5, 0.2] &  5000 \\
       $h_{01}$  & [-4.5, -1.2, 1.2]  &  [1.2, 1.5, 2.0]  & [0.3, 0.5, 0.2] &  10000 \\
       $h_{10}$  & [-1.8, 1.5, 6.0]  &  [1.2, 1.3, 2.0]   & [0.2, 0.5, 0.3] & 15000 \\
       $h_{11}$  &[-1.1, 2.3, 7.0]  & [1.2, 1.5, 2.0]    & [0.2, 0.4, 0.4] & 10000
    \end{tabular}
    \caption{Configuration of each synthetic data distribution $h_{ya}$.}
    \label{tab:syn}
\end{table}


\autoref{fig:kde} illustrates histograms of logit $h$ distributions of synthetic data and their KDE results in colored lines.
The top plot is about positive samples \ie $h_{11}$ and $h_{10}$, and the bottom plot is about positive samples \ie $h_{00}$ and $h_{01}$ respectively. We could observe that KDE accurately estimated the density function $h$ that cannot be fitted with parametric distribution.

Moreover, we conduct experiments to validate GSTAR can achieve the proposed goal.
Given 4 probability distributions and number of samples for each group as in \autoref{tab:syn}, we divide the dataset into training (70\%), validation (15\%), and testing (15\%) set. 
We train GSTAR on training set and find the best $\boldsymbol{\theta}$ by selecting one that has minimum validation loss and report the result on testing set.

In Figure \ref{fig:kde_perf} and \ref{fig:kde_theta}, we quantitatively evaluate GSTAR with KDE method with different $\lambda$ values on fairness constraint $\LL_{fair}$. In \autoref{fig:kde_perf}, color of the lines are performance and fairness measure as described in the legend. Dotted lines indicate baseline ($\boldsymbol{\theta} = (0,0)$) and solid lines indicate the measures of GSTAR. 
Note that GSTAR improve target fairness significantly with small lose of accuracy. In DP+EOd constraint, we even achieve almost perfect equal opportunity \ie Eq. Opp $\approx$ 0 with high enough $\lambda$ values.
\bibliographystyle{aaai22}
\bibliography{egbib}

\end{document}